\documentclass{article} % For LaTeX2e
\usepackage{iclr2025_conference,times}

\usepackage{amsmath,amsfonts,bm}

\def\eqref#1{equation~\ref{#1}}
\def\1{\bm{1}}

\DeclareMathAlphabet{\mathsfit}{\encodingdefault}{\sfdefault}{m}{sl}
\SetMathAlphabet{\mathsfit}{bold}{\encodingdefault}{\sfdefault}{bx}{n}

\usepackage{hyperref}
\usepackage{marvosym}
\usepackage{url}
\usepackage{graphicx}
\usepackage{subfigure}
\usepackage{booktabs}
\usepackage{caption}
\title{Hallo2: Long-Duration and High-Resolution Audio-Driven Portrait Image Animation}

\iclrfinalcopy
\author{
Jiahao Cui$^{1\ast}$, 
Hui Li$^{1\ast}$,
Yao Yao$^{3}$,
Hao Zhu$^{3}$,
Hanlin Shang$^{1}$,
Kaihui Cheng$^{1}$,
Hang Zhou$^{2}$\\
\;\textbf{Siyu Zhu}$^{1\href{siyuzhu@fudan.edu.cn}{\textrm{\Letter}}}$,
\textbf{Jingdong Wang}$^{2}$\\
$^{1}$Fudan University\;\;
$^{2}$Baidu Inc.\;\;
$^{3}$Nanjing University
}

\begin{document}

\maketitle

\begin{abstract}
Recent advances in latent diffusion-based generative models for portrait image animation, such as Hallo, have achieved impressive results in short-duration video synthesis. 
In this paper, we present updates to Hallo, introducing several design enhancements to extend its capabilities.
First, we extend the method to produce long-duration videos. 
To address substantial challenges such as appearance drift and temporal artifacts, we investigate augmentation strategies within the image space of conditional motion frames. Specifically, we introduce a patch-drop technique augmented with Gaussian noise to enhance visual consistency and temporal coherence over long duration.
Second, we achieve 4K resolution portrait video generation. To accomplish this, we implement vector quantization of latent codes and apply temporal alignment techniques to maintain coherence across the temporal dimension. 
By integrating a high-quality decoder, we realize visual synthesis at 4K resolution.
Third, we incorporate adjustable semantic textual labels for portrait expressions as conditional inputs. 
This extends beyond traditional audio cues to improve controllability and increase the diversity of the generated content.
To the best of our knowledge, Hallo2, proposed in this paper, is the first method to achieve 4K resolution and generate hour-long, audio-driven portrait image animations enhanced with textual prompts.
We have conducted extensive experiments to evaluate our method on publicly available datasets, including HDTF, CelebV, and our introduced ``Wild'' dataset. 
The experimental results demonstrate that our approach achieves state-of-the-art performance in long-duration portrait video animation, successfully generating rich and controllable content at 4K resolution for duration extending up to tens of minutes.
Project page: \url{https://fudan-generative-vision.github.io/hallo2}
\end{abstract}

\section{Introduction}
Portrait image animation—the process of creating animated videos from a reference portrait using various input signals such as audio~\cite{prajwal2020lip,tian2024emo,xu2024hallo,zhang2023sadtalker}, facial landmarks~\cite{wei2024aniportrait,chen2024echomimic}, or textual descriptions~\cite{xu2024vasa}—is a rapidly evolving field with significant potential across multiple domains. 
These domains include high-quality film and animation production, the development of virtual assistants, personalized customer service solutions, interactive educational content creation, and realistic character animation in the gaming industry. 
Consequently, the capability to generate long-duration, high-resolution, audio-driven portrait animations, particularly those assisted by textual prompts, is crucial for these applications. Recent technological advancements, notably in latent diffusion models, have significantly advanced this field.
% The Contrastive Language-Image Pretraining (CLIP) encoder has bridged the gap between vision and language, serving as a foundational component in diffusion models for conditional image generation. 
% Furthermore, latent stable diffusion techniques have markedly improved generation efficiency by producing high-quality images in latent space. 
% This approach facilitates the creation of detailed and visually appealing images while allowing precise control over generated features through the manipulation of latent variables. 
% In the realm of denoiser design, architectures such as UNet effectively recover image details using encoder-decoder structures with multi-scale feature extraction capabilities. 
% Alternatively, the Diffusion Transformer (DiT) model employs self-attention mechanisms to capture global contextual information. 
% Collectively, these advancements have significantly contributed to the progress and application of denoising techniques within the framework of stable diffusion models.

Several methods utilizing latent diffusion models for portrait image animation have emerged in recent years. 
For instance, VASA-1~\cite{xu2024vasa} employs the DiT model~\cite{Peebles2022DiT} as a denoiser in the diffusion process, converting a single static image and an audio segment into realistic conversational facial animations. 
Similarly, the EMO framework~\cite{tian2024emo} represents the first end-to-end system capable of generating animations with high expressiveness and realism, seamless frame transitions, and identity preservation using a U-Net-based diffusion model~\cite{blattmann2023stable} with only a single reference image and audio input. 
Other significant advancements in this domain include AniPortrait~\cite{wei2024aniportrait}, EchoMimic~\cite{chen2024echomimic}, V-Express~\cite{wang2024v}, Loopy~\cite{jiang2024loopy}, and CyberHost~\cite{lin2024cyberhost}, each contributing to enhanced capabilities and applications of portrait image animation. 
Hallo~\cite{xu2024hallo}, another notable contribution, introduces hierarchical audio-driven visual synthesis, building upon previous research to achieve facial expression generation, head pose control, and personalized animation customization. 
In this paper, we present updates to Hallo~\cite{xu2024hallo} by introducing several design enhancements to extend its capabilities.

\begin{figure*}[t]
\centering
\subfigure[Parallel Generation]{
\begin{minipage}[b]{0.47\textwidth}
\includegraphics[width=1.0\textwidth]{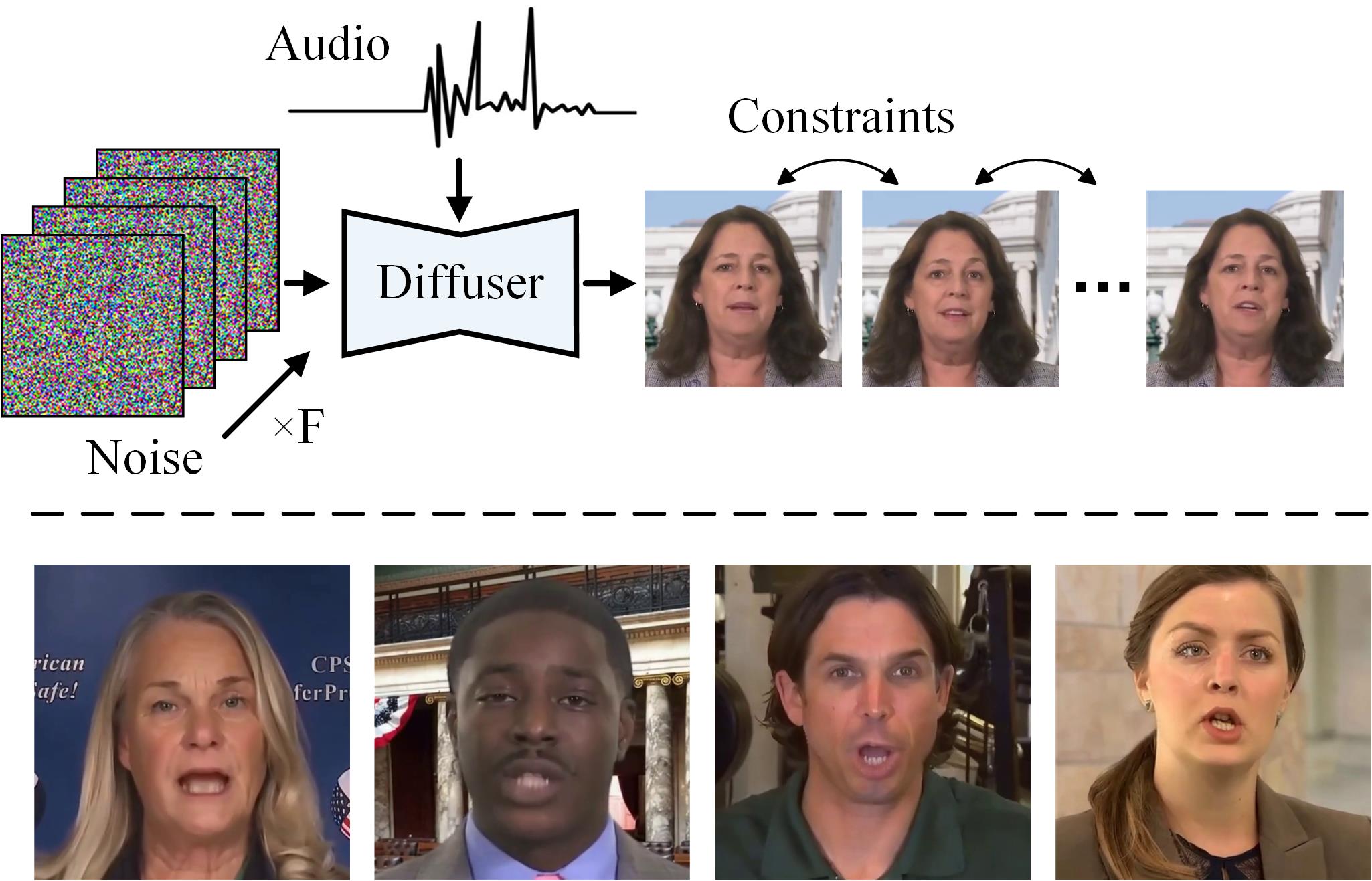}
\label{fig:parallelgeneration}
\vspace{-2mm}
\end{minipage}
}
\hspace{1mm}
\subfigure[Incremental Generation]{
\begin{minipage}[b]{0.47\textwidth}
\includegraphics[width=1.0\textwidth]{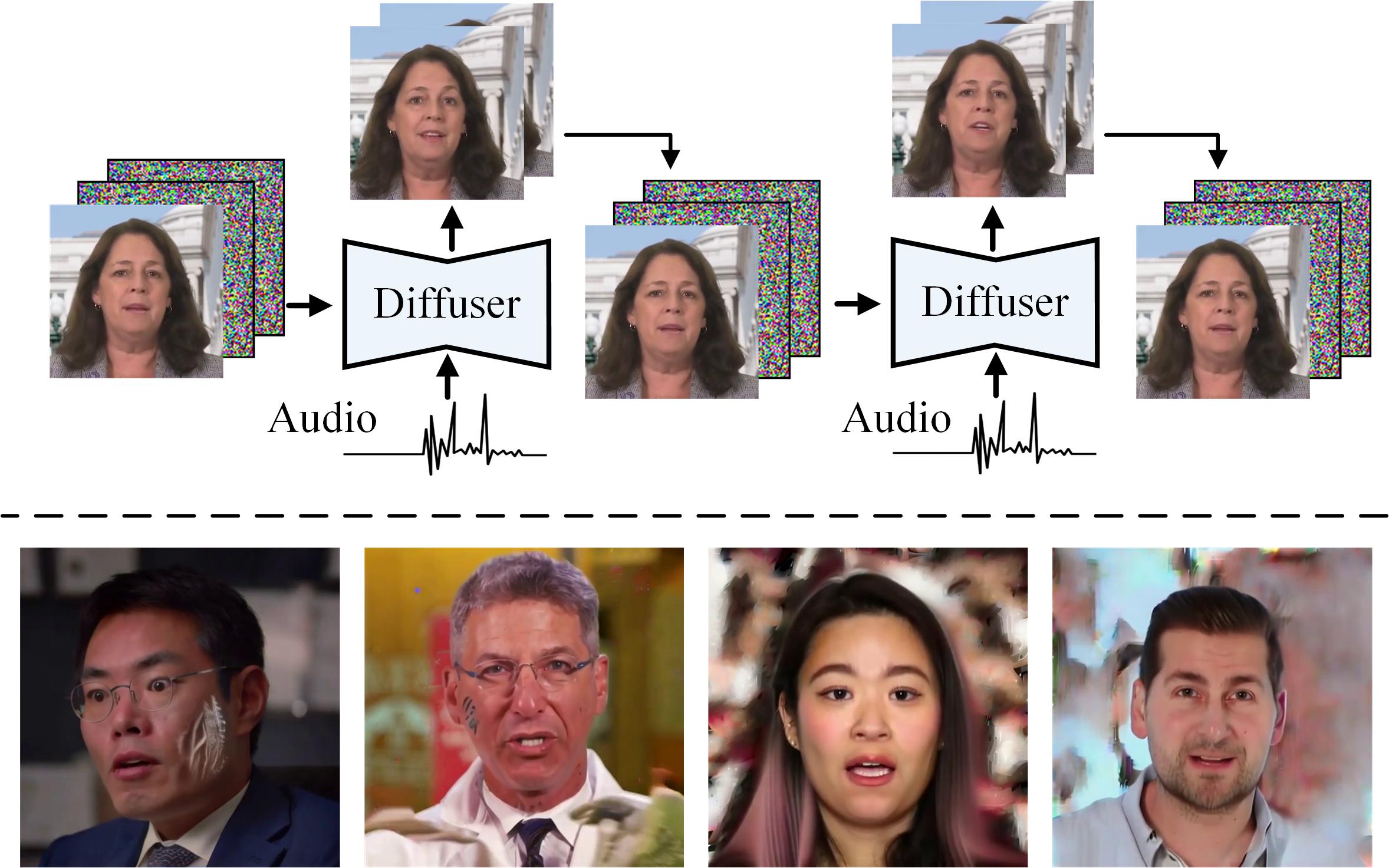}
\label{fig:incrementalgeneration}
\vspace{-2mm}
\end{minipage}
}
\vspace{-2mm}
\caption{
Comparison of parallel and incremental diffusion-based generative models for long-term portrait image animation. 
(a) The parallel generation approach may lead to blurriness and distorted expressions due to inter-frame continuity constraints. 
(b) The incremental generation method is susceptible to error accumulation in both facial features and backgrounds.
}
\vspace{-4mm}
\label{fig:generation}
\end{figure*}

Firstly, we extend Hallo from generating brief, second-long portrait animations to supporting duration of up to tens of minutes. 
As illustrated in Figure~\ref{fig:generation}, two primary approaches are commonly employed for long-term video generation. 
The first approach involves generating audio-driven video clips in parallel, guided by control signals, and then applying appearance and motion constraints between adjacent frames of these clips~\cite{wei2024aniportrait,chen2024echomimic}. 
A significant limitation of this method is the necessity to maintain minimal differences in appearance and motion across generated clips, which hampers substantial variations in lip movements, facial expressions, and poses, often resulting in blurriness and distorted expressions and postures due to the enforced continuity constraints.
The second approach incrementally generates new video content by leveraging preceding frames as conditional information~\cite{xu2024hallo,tian2024emo,wang2021audio2head}.
While this allows for continuous motion, it is prone to error accumulation.
Distortions, deformations relative to the reference image, noise artifacts, or motion inconsistencies in preceding frames can propagate to subsequent frames, degrading the overall video quality.
\begin{figure}[!t]
    \centering
    \includegraphics[width=0.98\linewidth]{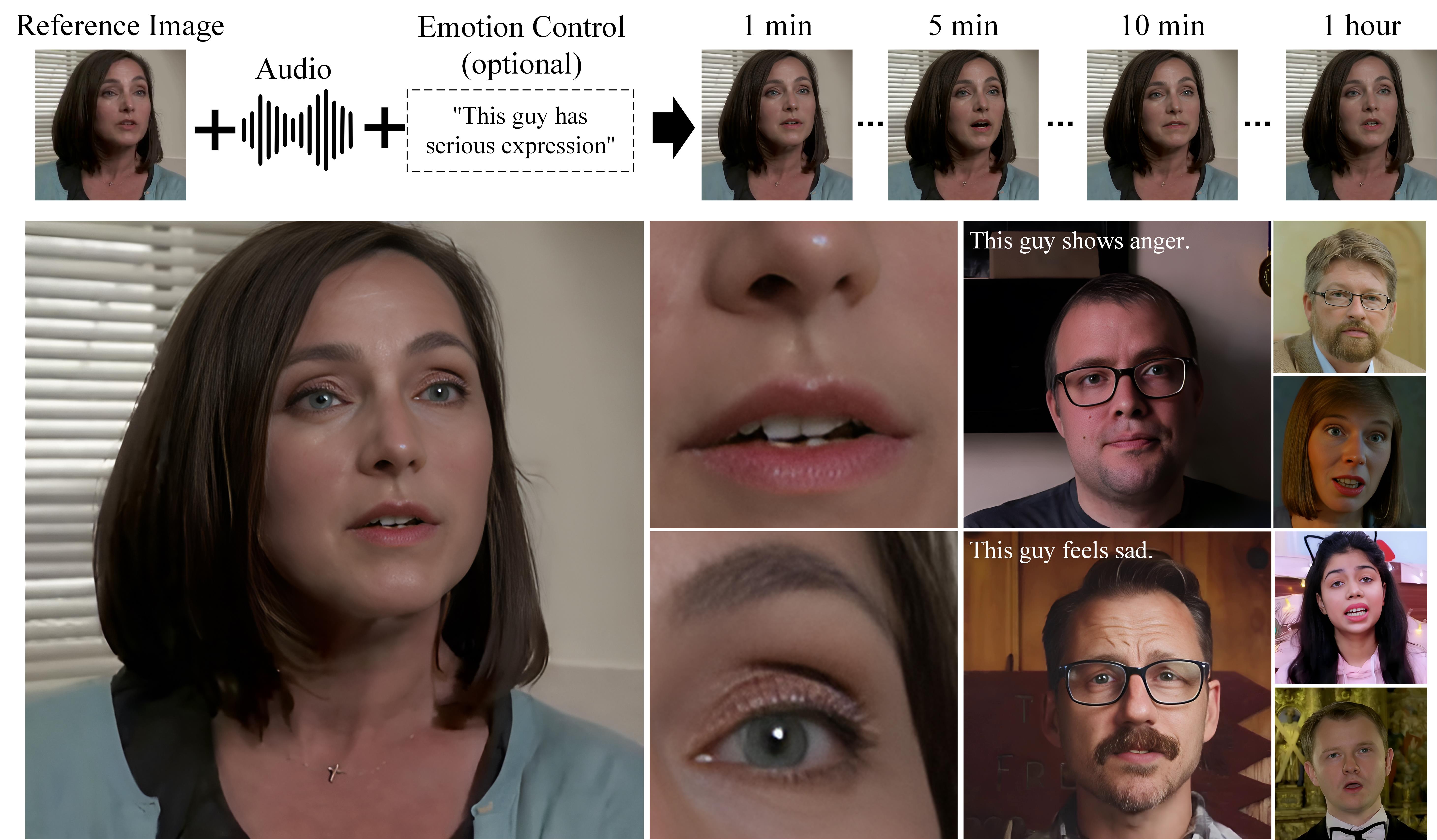}%teaser.png
    \vspace{-2mm}
    \caption{Demonstration of the proposed approach. 
    This approach processes a single reference image alongside an audio input lasting several minutes. 
Additionally, optional textual prompts may be introduced at various intervals to modulate and refine the expressions of the portrait. The resulting output is a high-resolution 4K video that synchronizes with the audio and is influenced by the optional expression prompts, ensuring continuity throughout the extended duration of the video.}
    \label{fig:teaser}
    \vspace{-4mm}
\end{figure}

To achieves high expressiveness, realism, and rich motion dynamics, we follow the second approach.
Our method primarily derives the appearance from the reference image, utilizing preceding generated frames solely to convey motion dynamics—including lip movements, facial expressions, and poses. 
To prevent contamination of appearance information from preceding frames, we implement a patch-drop data augmentation technique that introduces controlled corruption to the appearance information in the conditional frames while preserving motion characteristics. 
This approach encourages that the appearance is predominantly sourced from the reference portrait image, maintaining robust identity consistency throughout the animation and enabling long videos with continuous motion. 
Additionally, to enhance resilience against appearance contamination, we incorporate Gaussian noise as an additional data augmentation technique applied to the conditional frames, further reinforcing fidelity to the reference image while effectively utilizing motion information.

Secondly, to achieve 4K video resolution, we extend the Vector Quantized Generative Adversarial Network (VQGAN)~\cite{esser2021taming} discrete codebook space method for code sequence prediction tasks into the temporal dimension. 
By incorporating temporal alignment into the code sequence prediction network, we achieve smooth transitions in the predicted code sequences of the generated video. 
Upon applying the high-quality decoder, the strong consistency in both appearance and motion allows our method to enhance the temporal coherence of high-resolution details.

Thirdly, to enhance the semantic control of long-term portrait video generation, we introduce adjustable semantic textual prompt for portrait expressions as conditional inputs alongside audio signals. 
By injecting textual prompts at various time intervals, our method can help to adjust facial expressions and head poses, thereby rendering the animations more lifelike and expressive.

To evaluate the effectiveness of our proposed method, we conducted comprehensive experiments on publicly available datasets, including HDTF, CelebV, and our introduced ``Wild'' dataset. 
To the best of our knowledge, our approach is the first to achieve 4K resolution in portrait image animation for duration extending up to ten minutes or even several hours. 
Furthermore, by incorporating adjustable textual prompts that enable precise control over facial features during the generation process, our method ensures high levels of realism and diversity in the generated animations.

\section{Related Work}

% Our research builds upon recent advancements in video diffusion models, portrait image animation, and long-term high-resolution video generation. 
% This section reviews pertinent literature in these domains to position our work within the existing body of research.

\textbf{Video Diffusion Models.}
Diffusion-based models have demonstrated remarkable capabilities in generating high-quality and realistic videos from textual and image inputs~\cite{hu2023animate,zhu2024champ,zhang2024tora}. 
Stable Video Diffusion~\cite{blattmann2023stable} emphasizes latent video diffusion approaches, utilizing pretraining, fine-tuning, and curated datasets to enhance video quality. 
Make-A-Video~\cite{singer2022make} leverages text-to-image synthesis techniques to optimize text-to-video generation without requiring paired data. 
MagicVideo~\cite{zhou2022magicvideo} introduces an efficient framework with a novel 3D U-Net design, reducing computational costs. AnimateDiff~\cite{guo2023animatediff} enables animation of personalized text-to-image models via a plug-and-play motion module.
Further contributions, such as VideoComposer~\cite{wang2023videocomposer} and VideoCrafter~\cite{chen2023videocrafter1}, emphasize controllability and quality in video generation. 
VideoComposer integrates motion vectors for dynamic guidance, while VideoCrafter offers open-source models. CogVideoX~\cite{yang2024cogvideox} enhances text-video alignment through expert transformers, and MagicTime~\cite{yuan2024magictime} addresses the encoding of physical knowledge with a metamorphic time-lapse model.
Building upon these advancements, our approach adopts superior pretrained diffusion models tailored specifically for portrait image animation, focusing on long-duration and high-resolution synthesis.
\begin{figure}[!t]
    \centering
    \includegraphics[width=1.0\linewidth]{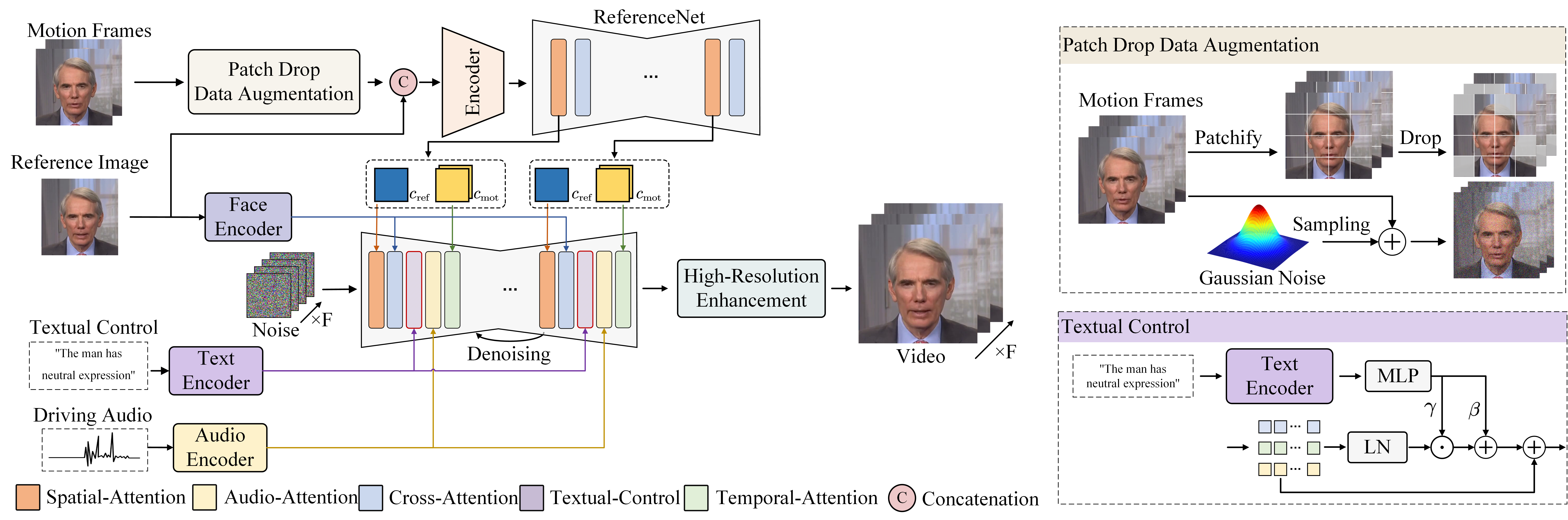}
    \vspace{-4mm}
    \caption{The framework of the proposed approach.
    The details of the proposed patch drop data augmentation and textual prompt control are shown on the right side.
    $c_{\mathtt{ref}}$ and $c_{\mathtt{mot}}$ refer to the feature of reference image and motion frames.}
    \label{fig:overview}
    \vspace{-5mm}
\end{figure}

\textbf{Portrait Image Animation.}
Significant progress has been made in audio-driven talking head generation and portrait image animation, emphasizing realism and synchronization with audio inputs. 
LipSyncExpert~\cite{prajwal2020lip} improved lip-sync accuracy using discriminators and novel evaluation benchmarks. 
Subsequent methods like SadTalker~\cite{zhang2023sadtalker} and VividTalk~\cite{sun2023vividtalk} incorporated 3D motion modeling and head pose generation to enhance expressiveness and temporal synchronization.
Diffusion-based techniques have further advanced the field. DiffTalk~\cite{shen2023difftalk} and DreamTalk~\cite{ma2023dreamtalk} improved video quality while maintaining synchronization across diverse identities. 
VASA-1~\cite{xu2024vasa} and AniTalker~\cite{liu2024anitalker} integrated nuanced facial expressions and universal motion representations, resulting in lifelike and synchronous animations. 
AniPortrait~\cite{wei2024aniportrait}, EchoMimic~\cite{chen2024echomimic}, V-Express~\cite{wang2024v}, Loopy~\cite{jiang2024loopy}, CyberHost~\cite{lin2024cyberhost}, and EMO~\cite{tian2024emo} have contributed to enhanced capabilities, focusing on expressiveness, realism, and identity preservation.
Despite these advancements, generating long-duration, high-resolution portrait videos with consistent visual quality and temporal coherence remains a challenge. 
Our method builds upon Hallo~\cite{xu2024hallo} to address this gap by achieving realistic, high-resolution motion dynamics in long-term portrait image animations.

\textbf{Long-Term and High-Resolution Video Generation.}
Recent advances in video diffusion models have significantly enhanced the generation of long-duration, high-resolution videos. 
Frameworks like Flexible Diffusion Modeling~\cite{harvey2022flexible} and Gen-L-Video~\cite{harvey2022flexible} improve temporal coherence and enable text-driven video generation without additional training. 
Methods such as SEINE~\cite{chen2023seine} and StoryDiffusion~\cite{zhou2024storydiffusion} introduce generative transitions and semantic motion predictors for smooth scene changes and visual storytelling. 
Approaches like StreamingT2V~\cite{henschel2024streamingt2v} and MovieDreamer~\cite{zhao2024moviedreamer} use autoregressive strategies and diffusion rendering for extended narrative videos with seamless transitions. 
Video-Infinity\cite{tan2024video} optimizes long video synthesis through distributed inference, while FreeLong~\cite{lu2024freelong} integrates global and local video features without training for consistency.
In this paper, we employ patch-drop and Gaussian noise augmentation to enable long-duration portrait image animation.

Discrete prior representations with learned dictionaries have proven effective for image restoration. 
VQ-VAE~\cite{razavi2019generating} enhances VAEs by introducing discrete latent spaces via vector quantization, addressing posterior collapse, and enabling high-quality image, video, and speech generation. 
Building on this, VQ-GAN~\cite{lee2022autoregressive} combines CNNs and Transformers to create a context-rich vocabulary of image components, achieving state-of-the-art results in conditional image generation. 
CodeFormer~\cite{zhou2022towards} uses a learned discrete codebook for blind face restoration, employing a Transformer-based network for enhanced robustness against degradation.
This paper proposes vector quantization of latent codes with temporal alignment techniques to maintain high-resolution coherence temporally for 4K synthesis.
\section{Preliminaries}

\subsection{Latent Diffusion Models}

Latent Diffusion Models (LDMs), introduced by~\cite{rombach2022high}, represent a significant advancement in generative modeling by conducting diffusion and denoising processes within a compressed latent space rather than directly in the high-dimensional image space. 
This approach substantially reduces computational complexity while maintaining the quality of generated images.

Specifically, a pre-trained Variational Autoencoder (VAE)~\cite{kingma2013auto} is employed to encode input images into lower-dimensional latent representations. 
Given an input image $\mathbf{I}$, the encoder $\mathcal{E}(\cdot)$ maps it to a latent vector: $\mathbf{z}_0 = \mathcal{E}(\mathbf{I})$.
A forward stochastic diffusion process~\cite{sohl2015deep,ho2020denoising,song2020score} is then applied to the latent vector $\mathbf{z}_0$, adding Gaussian noise over $T$ time steps to produce a sequence of noisy latent variables $\{\mathbf{z}_t\}_{t=1}^T$. 
The process is defined by:
\begin{equation}
\label{eq:diffusion_forward}
\mathbf{z}_t = \sqrt{\bar{\alpha}_t} \, \mathbf{z}_0 + \sqrt{1 - \bar{\alpha}_t} \, \boldsymbol{\epsilon},
\boldsymbol{\epsilon} \sim \mathcal{N}(\mathbf{0}, \mathbf{I}),
\end{equation}
where $t \in \{1, 2, \ldots, T\}$ denotes the diffusion steps, $\alpha_t = 1 - \beta_t$ with $\beta_t \in (0,1)$ being the variance schedule, and $\bar{\alpha}_t = \prod_{s=1}^t \alpha_s$ is the cumulative product of $\alpha_t$. 
As $t$ approaches $T$, the distribution of $\mathbf{z}_T$ converges to a standard normal distribution $\mathcal{N}(\mathbf{0}, \mathbf{I})$ due to the accumulated noise.

The reverse diffusion process aims to reconstruct the original latent vector $\mathbf{z}_0$ by sequentially denoising $\mathbf{z}_T$. At each timestep $t$, a noise prediction network $\boldsymbol{\epsilon}_{\theta}$, typically parameterized using a U-Net architecture~\cite{ronneberger2015u}, estimates the noise component in $\mathbf{z}_t$ using optional conditioning information $\mathbf{c}$. The network is trained to minimize the expected mean squared error between the true noise $\boldsymbol{\epsilon}$ and the predicted noise $\boldsymbol{\epsilon}_{\theta}$:
\begin{equation}
\label{eq:training_objective}
\mathcal{L} = \mathbb{E}_{\mathbf{z}_0, \mathbf{c}, \boldsymbol{\epsilon}, t} \left[ \omega(t) \left\| \boldsymbol{\epsilon} - \boldsymbol{\epsilon}_{\theta}(\mathbf{z}_t, t, \mathbf{c}) \right\|_2^2 \right],
\end{equation}
where $\omega(t)$ is a weighting function that balances the loss contribution across different timesteps.

Once trained, the model can generate new samples by starting from a random Gaussian latent vector $\mathbf{z}_T \sim \mathcal{N}(\mathbf{0}, \mathbf{I})$ and iteratively applying the denoising process:
\begin{equation}
\mathbf{z}_{t-1} = \frac{1}{\sqrt{\alpha_t}} \left( \mathbf{z}_t - \frac{1 - \alpha_t}{\sqrt{1 - \bar{\alpha}_t}} \, \boldsymbol{\epsilon}_{\theta}(\mathbf{z}_t, t, \mathbf{c}) \right) + \sigma_t \, \mathbf{n}, \mathbf{n} \sim \mathcal{N}(\mathbf{0}, \mathbf{I}),
\end{equation}
for $t = T, T-1, \ldots, 1$, where $\sigma_t$ is the standard deviation of the noise added at step $t$. The final latent vector $\mathbf{z}_0$ is then decoded to reconstruct the image: $\mathbf{I} = \mathcal{D}(\mathbf{z}_0)$,
where $\mathcal{D}(\cdot)$ is the decoder of the  Variational Autoencoder (VAE).

\subsection{Incorporating Motion Conditions via Cross-Attention}
Incorporating conditioning information is crucial for controlling the generative process in latent diffusion models. Cross-attention mechanisms~\cite{vaswani2017attention} are employed to effectively integrate motion conditions into the model. The attention layers process both the noisy latent variables $\mathbf{z}_t$ and the embedded motion conditions $\mathbf{c}$ to guide the denoising process.
The cross-attention operation is formulated as:
\begin{equation}
\label{eq:cross_attention}
\text{CrossAttn}(\mathbf{z}_t, \mathbf{c}) = \text{softmax}\left(\mathbf{Q} \mathbf{K}^\top/\sqrt{d_k}\right) \mathbf{V},
\end{equation}
where $\mathbf{Q} = \mathbf{W}_Q \mathbf{z}_t$, $\mathbf{K} = \mathbf{W}_K \mathbf{c}$ and $\mathbf{V} = \mathbf{W}_V \mathbf{c}$ are the queries;
$\mathbf{W}_Q$, $\mathbf{W}_K$, and $\mathbf{W}_V$ are learnable projection matrices;
and $d_k$ is the dimensionality of the keys. 
The softmax function ensures that the attention weights sum to one, focusing on the most relevant components of the conditioning information.
By integrating cross-attention into the denoising network, the model dynamically adjusts its focus based on the current latent state and the provided conditions. This mechanism enables the generation of images that are coherent with the conditioning inputs, enhancing the expressiveness and realism of the animated portraits.

In our work, the motion conditions $\mathbf{c}$ include the reference image embedding $\mathbf{c}_{\text{image}}$, audio features $\mathbf{c}_{\text{audio}}$, and textual embeddings $\mathbf{c}_{\text{text}}$ obtained via Contrastive Language-Image Pretraining (CLIP)~\cite{radford2021learning}. The combination of these modalities allows for nuanced control over facial expressions, lip movements, and head poses in the generated animations.
\section{Method}

\begin{figure}[!t]
    \centering
    \includegraphics[width=0.9\linewidth]{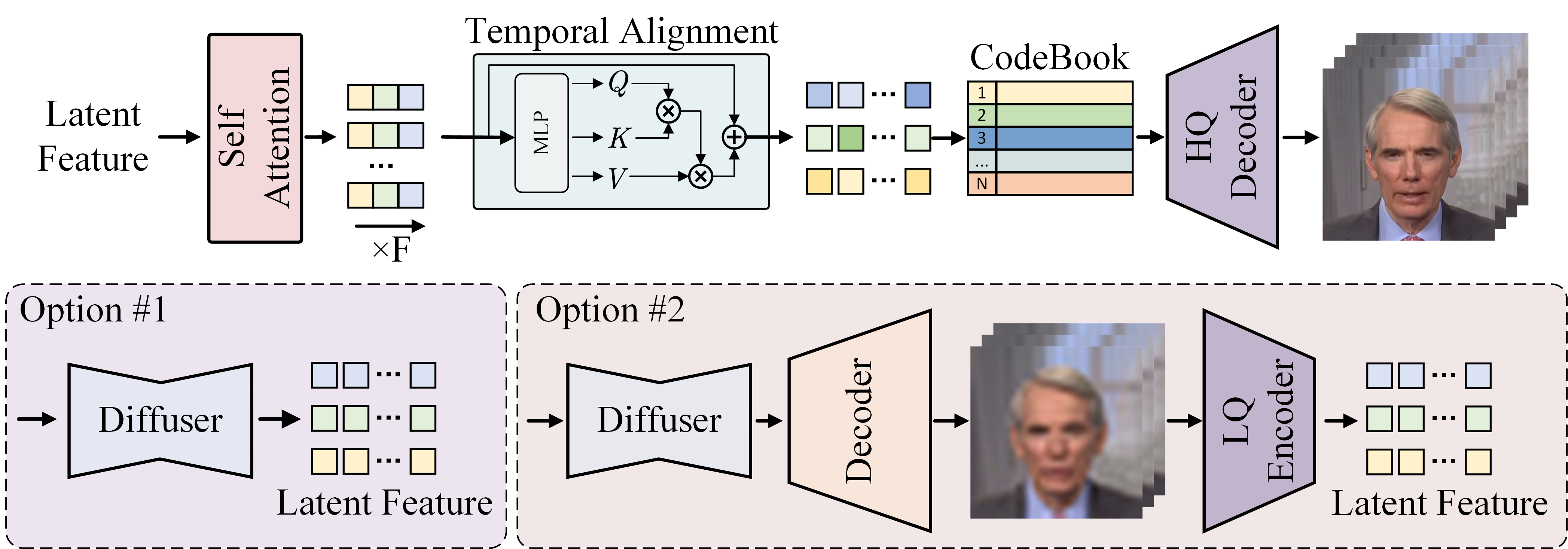}
    \vspace{-2mm}
    \caption{The illustration of the proposed high-resolution enhancement module.
    Two alternative designs for extracting input latent features are demonstrated.}
    \label{fig:high-resolution}
    \vspace{-4mm}
\end{figure}

In this section, we introduce an extended technique for portrait image animation that effectively addresses the challenges of generating long-duration, high-resolution videos with intricate motion dynamics, as well as enabling audio-driven and textually prompted control.
Our proposed method derives the subject's appearance primarily from a single reference image while utilizing preceding generated frames as conditional inputs to capture motion information.
To preserve appearance details of the reference image and prevent contamination from preceding frames, we introduce a patch drop data augmentation technique combined with Gaussian noise injection (see Section~\ref{subsec:drop}).
Additionally, we extend the VQGAN discrete codebook prediction into the temporal domain, facilitating high-resolution video generation and enhancing temporal coherence (see Section~\ref{subsec:high_resolution}).
Furthermore, we integrate textual conditions alongside audio signals to enable diverse control over facial expressions and motions during long-term video generation (see Section~\ref{subsec:text_control}).
Finally, we detail the network structure along with the training and inference strategies in Section~\ref{subsec:network}.

\subsection{Long-Duration Animation}\label{subsec:drop}

\textbf{Patch-Drop Augmentation.}
To generate long-duration portrait videos that maintain consistent appearance while exhibiting rich motion dynamics, we introduce a patch drop data augmentation technique applied to the conditioning frames. 
The core idea is to corrupt the appearance information in preceding frames while preserving their motion cues, thereby ensuring that the model relies primarily on the reference image for appearance features and utilizes preceding frames to capture temporal dynamics.

Let $\mathbf{I}_{\text{ref}}$ denote the reference image, and let $\{\mathbf{I}_{t-1}, \mathbf{I}_{t-2}, \dotsc, \mathbf{I}_{t-N}\}$ represent the preceding $N$ generated frames at time steps $t-1$ to $t-N$. 
To mitigate the influence of appearance information from preceding frames, we apply a patch drop augmentation to each frame $\mathbf{I}_{t-i}$, $\text{for } i = 1, 2, \dotsc, N$.  
Specifically, each frame is partitioned into $K$ non-overlapping patches of size $p \times p$, yielding $\{\mathbf{I}_{t-i}^{(k)}\}_{k=1}^K$, where $k$ indexes the patches. For each patch, a binary mask $M_{t-i}^{(k)}$ is generated as follows:
\begin{equation}
M_{t-i}^{(k)} = 
\begin{cases} 
1 & \text{if } \xi^{(k)} \geq r \\ 
0 & \text{if } \xi^{(k)} < r
\end{cases}
\end{equation}
Here $\xi^{(k)} \sim \mathcal{U}(0, 1)$ is a uniformly distributed random variable, and $r \in [0,1]$ is the patch drop rate controlling the probability of retaining each patch.

The augmented frame $\tilde{\mathbf{I}}_{t-i}$ is then constructed by applying the masks to the corresponding patches:
\begin{equation}
\tilde{\mathbf{I}}_{t-i}^{(k)} = M_{t-i}^{(k)} \cdot \mathbf{I}_{t-i}^{(k)},\; \text{for } k = 1, 2, \dotsc, K.
\end{equation}
This random omission of patches effectively disrupts detailed appearance information while preserving the coarse spatial structure necessary for modeling motion dynamics.

\textbf{Gaussian Noise Augmentation.}
During the incremental generation process, previously generated video frames may introduce contamination in both appearance and dynamics, such as noise in facial regions and the background, or subtle distortions in lip movements and facial expressions.
As this process continues, these contaminations can propagate to subsequent frames, leading to the gradual accumulation and amplification of artifacts. 
To mitigate this issue, we incorporate Gaussian noise into the motion frames, enhancing the denoiser's ability in the latent space to recover from contaminations in appearance and dynamics.
Specifically, we introduce Gaussian noise to the augmented latent representations:
\begin{equation}
\hat{\mathbf{z}}_{t-i} = \tilde{\mathbf{z}}_{t-i} + \boldsymbol{\eta}_{t-i}, 
\boldsymbol{\eta}_{t-i} \sim \mathcal{N}(\mathbf{0}, \sigma^{2} \mathbf{I}),
\end{equation}
where \(\sigma\) controls the noise level, and \(\mathbf{I}\) denotes the identity matrix. 
The corrupted latent representations \(\{\hat{\mathbf{z}}_{t-1}, \hat{\mathbf{z}}_{t-2}, \dotsc, \hat{\mathbf{z}}_{t-N}\}\) are then used as motion condition inputs to the diffusion model.

These noise-augmented motion frames are incorporated into the diffusion process via cross-attention mechanisms within the denoising U-Net. At each denoising step \(t\), the model predicts the noise component \(\boldsymbol{\epsilon}_{\theta}(\mathbf{z}_t, t, \mathbf{c})\), where \(\mathbf{z}_t\) is the current noisy latent, and \(\mathbf{c}\) represents the set of conditioning inputs:
\begin{equation}
\boldsymbol{\epsilon}_{\theta}(\mathbf{z}_t, t, \mathbf{c}) = \boldsymbol{\epsilon}_{\theta}\left(\mathbf{z}_t, t, \mathbf{z}_{\text{ref}}, \{\hat{\mathbf{z}}_{t-i}\}, \mathbf{c}_{\text{audio}}, \mathbf{c}_{\text{text}}\right).
\end{equation}
Here, \(\mathbf{z}_{\text{ref}} = \mathcal{E}(\mathbf{I}_{\text{ref}})\) is the latent representation of the reference image, and \(\mathbf{c}_{\text{audio}}\), \(\mathbf{c}_{\text{text}}\) are the encoded audio features and textual embeddings, respectively. 
By leveraging the noise-augmented motion frames, the model effectively captures temporal dynamics while mitigating the influence of accumulated artifacts. 
This approach encourages that the subject's appearance remains stable, derived from the reference image, throughout the generated video sequence. 

% Collectively, these two augmentation strategies enhance the robustness of the denoiser against propagated contamination, facilitating the generation of long-duration videos characterized by consistent appearance and realistic, coherent variations in motion dynamics, guided by audio and textual inputs.

\subsection{High-Resolution Enhancement}\label{subsec:high_resolution}
To enhance temporal coherence in high-resolution video generation, we adopt a codebook prediction approach~\cite{zhou2022codeformer}, incorporating an introduced temporal alignment mechanism.
% In this framework, we fix the encoder $\mathcal{E}$, decoder $\mathcal{D}$, and codebook $C$, while introducing a Transformer module that incorporates temporal alignment layers into the super-resolution network.

Given the generated video frames, we first encode them using a fixed encoder $\mathcal{E}$ to obtain latent representations $\mathbf{z} \in \mathbb{R}^{N \times H \times W \times C}$, where $N$ denotes the number of frames, while $H$, $W$, and $C$ represent the height, width, and number of channels, respectively. 
Each Transformer block comprises a spatial self-attention layer followed by a temporal alignment layer. The operations of the spatial self-attention layer are defined as follows.
Let $\mathbf{W}_Q$, $\mathbf{W}_K$, and $\mathbf{W}_V$ be learnable projection matrices. Given the input $\mathbf{z}$ to this Transformer block, we compute the queries, keys, and values as follows:
\begin{equation}
\mathbf{Q}_{\text{self}} = \mathbf{W}_Q \mathbf{z},\;\;
\mathbf{K}_{\text{self}} = \mathbf{W}_K \mathbf{z},\;\;
\mathbf{V}_{\text{self}} = \mathbf{W}_V \mathbf{z}.
\end{equation}
Subsequently, the output of the spatial self-attention layer, denoted as $\mathbf{X}_{\text{self}}$, is computed using the softmax function:
\begin{equation}
\mathbf{X}_{\text{self}} = \text{Softmax}\left({\mathbf{Q}_{\text{self}} \mathbf{K}_{\text{self}}^T}/{\sqrt{d_k}}\right) \mathbf{V}_{\text{self}} + \mathbf{z},
\end{equation}
where $d_k$ is the dimensionality of the keys.
Following this, the hidden state $\mathbf{X}_{\text{self}} \in \mathbb{R}^{N \times (H \cdot W) \times C}$ is reshaped into $\mathbf{X}_{\text{temp}} \in \mathbb{R}^{(H \cdot W) \times N \times C}$ to facilitate temporal attention across frames:
$\mathbf{X}_{\text{temp}} = \text{ReshapeToTemporal}(\mathbf{X}_{\text{self}})$.
In this context, let $\mathbf{W}'_Q$, $\mathbf{W}'_K$, and $\mathbf{W}'_V$ be additional learnable projection matrices. The queries, keys, and values for the temporal alignment layer are computed as follows:
\begin{equation}
\mathbf{Q}_{\text{temp}} = \mathbf{W}'_Q \mathbf{X}_{\text{temp}},\;\;
\mathbf{K}_{\text{temp}} = \mathbf{W}'_K \mathbf{X}_{\text{temp}},\;\;
\mathbf{V}_{\text{temp}} = \mathbf{W}'_V \mathbf{X}_{\text{temp}}.
\end{equation}
The output of the temporal attention mechanism, denoted as $\widetilde{\mathbf{X}}_{\text{temp}}$, is computed similarly:
\begin{equation}
\widetilde{\mathbf{X}}_{\text{temp}} = \text{Softmax}\left({\mathbf{Q}_{\text{temp}} \mathbf{K}_{\text{temp}}^T}/{\sqrt{d_k}}\right) \mathbf{V}_{\text{temp}} + \mathbf{X}_{\text{temp}}.
\end{equation}
Finally, $\widetilde{\mathbf{X}}_{\text{temp}}$ is reshaped back to the original dimensions of $\mathbf{z} \in \mathbb{R}^{N \times H \times W \times C}$:
\begin{equation}
\mathbf{z} = \text{ReshapeBack}(\widetilde{\mathbf{X}}_{\text{temp}}).
\end{equation}

As shown in Figure~\ref{fig:high-resolution}, we propose two implementations for extracting input latent features. 
The first approach directly utilizes latent features from the diffusion model for the super-resolution module, which, while simple, requires end-to-end training of the entire module. 
The second approach processes latent features through the diffusion model's decoder and then a low-quality decoder, necessitating only the training of a lightweight temporal alignment module. 
Given the sparsity of super-resolution video data, the second approach demonstrates superior performance under limited training conditions.

By integrating spatial and temporal attention mechanisms within the Transformer module, the network effectively captures intra-frame and inter-frame dependencies, enhancing both temporal consistency and visual fidelity in high-resolution video outputs.

% By integrating both spatial and temporal attention mechanisms within the Transformer module, the network effectively captures intra-frame and inter-frame dependencies, thereby enhancing temporal consistency and visual fidelity in the high-resolution video outputs. This dual attention framework ensures that the generated videos not only maintain high resolution but also exhibit coherent motion dynamics across frames, resulting in a more realistic and engaging visual experience.

\subsection{Textual Prompt Control}\label{subsec:text_control}

To enable precise modulation of facial expressions and motions based on textual instructions, we incorporate an adaptive layer normalization mechanism into the denoising U-Net architecture. Given a text prompt, a text embedding $\mathbf{e}_{\text{text}}$ is extracted using the CLIP text encoder~\cite{radford2021learning}. This embedding is processed through a zero-initialized multilayer perceptron (MLP) to produce scaling ($\gamma$) and shifting ($\beta$) parameters:
$
\gamma, \beta = \mathrm{MLP}(\mathbf{e}_{\text{text}}).
$

The adaptive layer normalization is applied between the cross-attention layer and the audio attention layer within the denoising U-Net. Specifically, the intermediate features $\mathbf{X}_{\text{cross}}$ from the cross-attention layer are adjusted as follows:
$
    \mathbf{X}_{\text{norm}} = \mathrm{LayerNorm}(\mathbf{X}_{\text{cross}}),
    \mathbf{X}_{\text{adapted}} = \gamma \odot \mathbf{X}_{\text{norm}} + \beta + \mathbf{X}_{\text{cross}},
$
where $\odot$ denotes element-wise multiplication. This adaptation conditions the denoising process on the textual input, enabling fine-grained control over the synthesized expressions and motions in the generated video frames.

\begin{figure}[!t]
    \begin{minipage}{0.52\textwidth}
        \centering
        \captionsetup{type=table}
        \resizebox{\textwidth}{!}{
        \begin{tabular}{c|c|c|c|c|c}
            \toprule
            \textbf{Method} & \textbf{FID$\downarrow$} & \textbf{FVD$\downarrow$} & \textbf{Sync-C$\uparrow$} & \textbf{Sync-D$\downarrow$} & \textbf{E-FID$\downarrow$} \\ 
            \midrule
            Audio2Head  &41.753  &246.041  &\textbf{8.051}  &\textbf{7.117}  &10.190 \\
            SadTalker  &21.924  &293.084  &7.399  &7.812  &6.881  \\
            % \midrule
            EchoMimic  &47.331  &532.733  &5.930  &9.143  &11.051  \\ 
            AniPortrait &26.241  &361.978  &3.912  &10.264  &11.253  \\ 
            % \midrule
            Hallo  &16.748  &366.066  &7.268  &7.714  &7.081  \\ 
            Ours &\textbf{16.616}  &\textbf{239.517}  &7.379  &7.697  &\textbf{6.702}  \\
            \midrule
            Real video & - & - &8.377 &6.809 & - \\ 
            \bottomrule
        \end{tabular}}
        \caption{The quantitative comparisons with existed portrait image animation approaches on the HDTF dataset. 
        Our evaluation focuses on generated videos with a duration of 4 minutes, maintaining consistent settings across subsequent quantitative experiments.}
    \label{tab:hdtf_results}
    \end{minipage} \hfill
     \begin{minipage}{0.44\textwidth}
    \centering
      \includegraphics[width=1.0\textwidth]{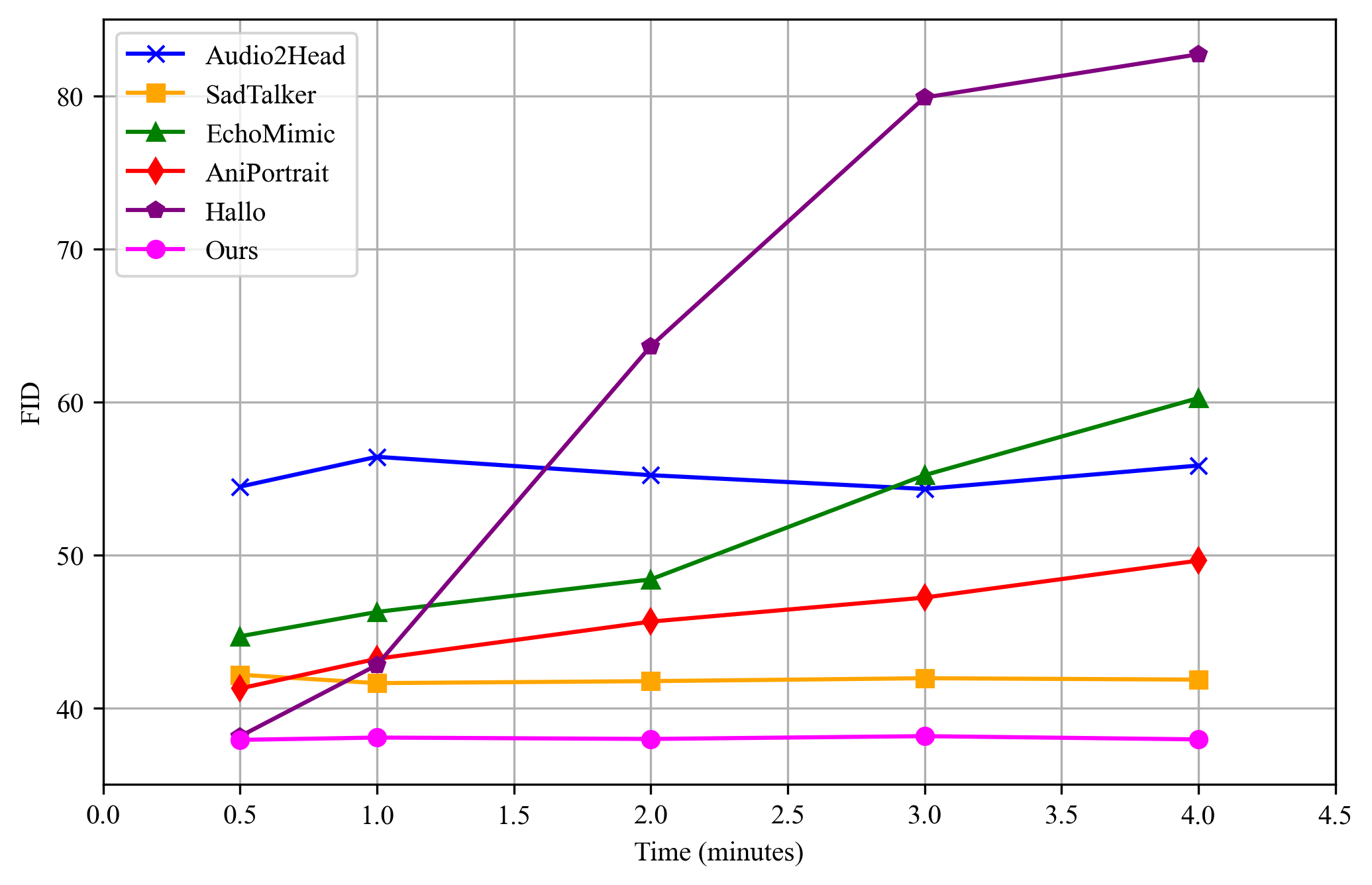}
      \vspace{-5mm}
      \caption{FID metrics of different methods as inference time increases.}\label{fig:fid_curve}
\end{minipage}
\end{figure}

\begin{figure}[!t]
    \centering
    \includegraphics[width=1.0\linewidth]{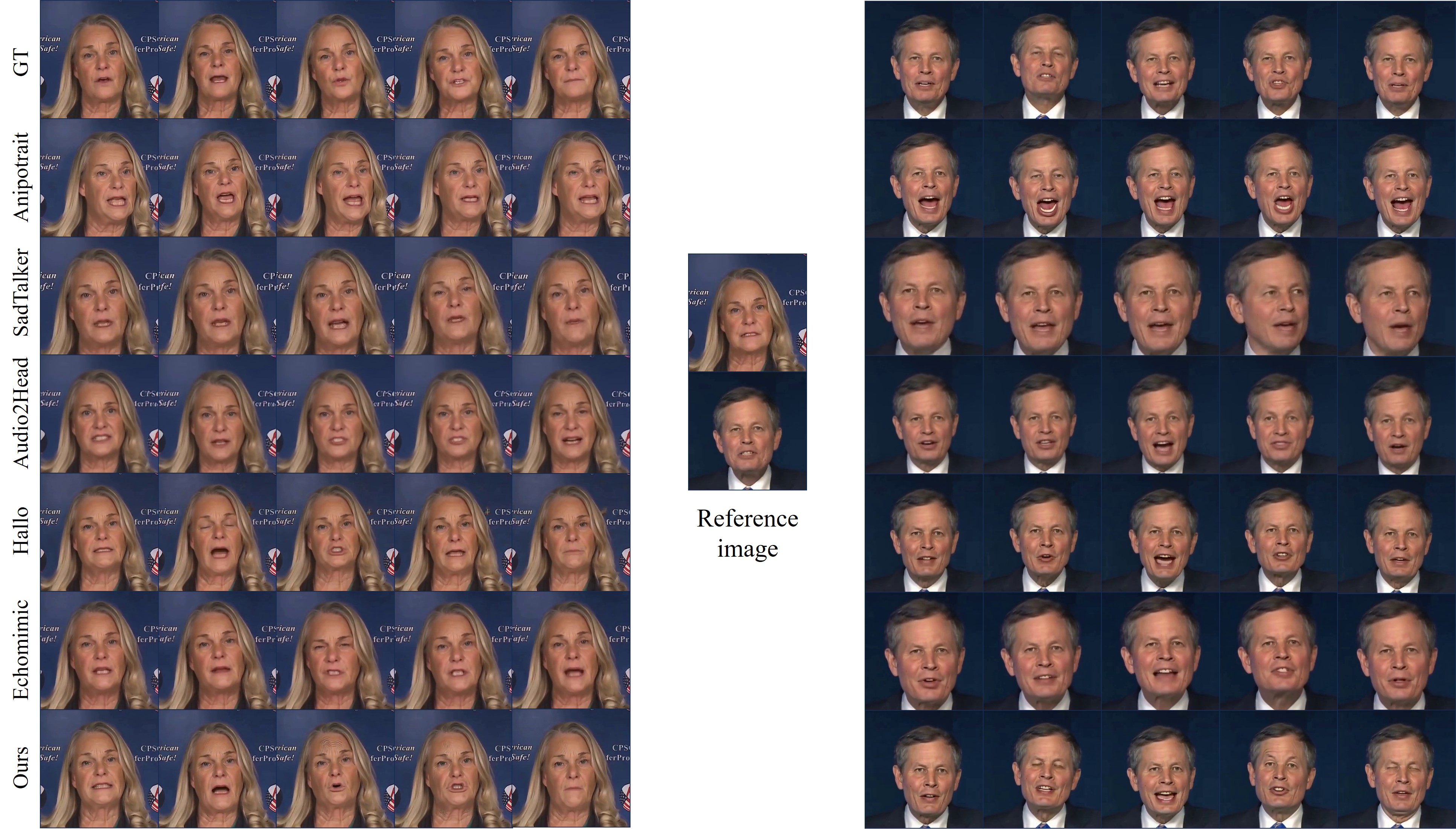}
    \vspace{-4mm}
    \caption{The qualitative comparison with exited approaches on HDTF data-set.}
    \label{fig:hdtf} % 创建标签以便引用
    \vspace{-4mm}
\end{figure}

\subsection{Network}\label{subsec:network}

\textbf{Network Architecture.}  
Figure~\ref{fig:overview} illustrates the proposed approach's architecture. 
The ReferenceNet embeds the reference image \(\mathbf{z}_\text{ref}\), capturing the visual appearance of both the portrait and the corresponding background. 
To model temporal dynamics while mitigating appearance contamination from preceding frames, the motion frames $\{\hat{\mathbf{z}}_{t-i}\}$ are subjected to patch dropping and Gaussian noise augmentation. Our extended framework utilizes a denoising U-Net architecture that processes noisy latent vectors \(\mathbf{z}_{t}\) at each diffusion timestep \(t\). 
The embedding of the input audio \(\mathbf{c}_{\text{audio}}\) is derived from a 12-layer wav2vec network~\cite{schneider2019wav2vec}, while the textual prompt embedding \(\mathbf{c}_{\text{text}}\) is obtained through CLIP~\cite{radford2021learning}.
By synthesizing these diverse conditioning inputs via cross-attention layers within the denoising U-Net~\cite{blattmann2023stable}, the model generates frames that maintain visual coherence with the reference image while dynamically exhibiting nuanced and expressive lip motions and facial expressions. 
Finally, the high-resolution enhancement module employs vector quantization of latent codes in conjunction with temporal alignment techniques to produce final videos at 4K resolution.

\textbf{Training.}
This study implements a two-stage training process aimed at optimizing distinct components of the overall framework.

In the initial stage, the model is trained to generate video frames using a reference image, input-driven audio, and a target video frame. 
During this phase, the parameters of the Variational Autoencoder (VAE) encoder and decoder, as well as those of the facial image encoder, are held constant. 
The optimization process focuses on the spatial cross-attention modules within both the ReferenceNet and the denoising U-Net, with the objective of enhancing the model's capabilities for portrait video generation. 
Specifically, a random image is selected from the input video clip to serve as the reference image, while adjacent frames are designated as target images for training purposes. 
Additionally, motion modules are introduced to improve the model's temporal coherence and smoothness.

In the second stage, patch drop and Gaussian noise augmentation techniques are applied to the motion frames to train the model for generating long-duration videos characterized by temporal coherence and smooth transitions. 
This stage refines the modeling of temporal dynamics by incorporating corrupted motion frames into the conditioning set, thereby enhancing the model's ability to capture motion continuity over extended sequences. 
Concurrently, textual prompts are utilized at this stage to facilitate precise modulation of facial expressions and motions based on textual instructions. 
For the super-resolution model, the parameters of the VAE encoder are optimized, with a focus on refining the weights responsible for codebook prediction. 
Temporal alignment is employed within the Transformer-based architecture to ensure consistency and high-quality outputs across frames, thereby enhancing temporal coherence in high-resolution details.

\textbf{Inference.}
During inference, the video generation network receives a single reference image, driving audio, an optional textual prompt, and motion frames augmented using patch dropping and Gaussian noise techniques as inputs. 
The network generates a video sequence that animates the reference image in accordance with the provided audio and textual prompt, synthesizing realistic lip movements and expressions synchronized with the audio output. 
Subsequently, the high-resolution enhancement module processes the generated video to produce high-resolution frames, thereby enhancing visual quality and fine facial details.
\section{Experiments}
\subsection{Experimental Setups}
\textbf{Implementation.}
All experiments were conducted on a GPU server equipped with 8 NVIDIA A100 GPUs. 
The training process was executed in two stages: the first stage comprised 30,000 steps with a step size of 4, targeting a video resolution of 512 × 512 pixels. 
The second stage involved 28,000 steps with a batch size of 4, initializing the motion module with weights from Animatediff. 
Approximately 160 hours of video data were utilized across both stages, with a learning rate set at 1e-5.
For the super-resolution component, training for temporal alignment was extended to 550,000 steps, leveraging initial weights from CodeFormer and a learning rate of 1e-4, using the VFHQ dataset as the super-resolution training data. 
Each instance in the second stage generated 16 video frames, integrating latents from the motion module with the first 4 ground truth frames, designated as motion frames. 
During inference, the output video resolution is increased to a maximum of 4096 × 4096 pixels.
% This structured approach ensured high fidelity and coherence in the generated animations throughout the training process.

\textbf{Datasets.}
To evaluate our proposed method, we employed several publicly available datasets, including HDTF, CelebV, and our introduced ``Wild'' dataset. 
The ``Wild'' dataset comprises 2019 clips, totaling approximately 155.9 hours of video content, featuring a diverse array of lip motions, facial expressions, and head poses. 
This extensive dataset provides a solid foundation for training and testing our portrait image animation framework, facilitating a comprehensive assessment of its ability to generate high-quality and expressive animations across various scenarios.

\textbf{Evaluation Metrics.} 
We employ several evaluation metrics to rigorously evaluate our portrait image animation framework. 
The Fréchet Inception Distance (FID) measures the statistical distance between generated and real images in feature space, with lower values indicating higher quality. 
The Fréchet Video Distance (FVD) extends this concept to video, assessing the similarity between generated and real videos, where lower values signify superior visual quality. 
The Sync-C metric gauges lip synchronization consistency with audio, with higher scores reflecting better alignment. 
Conversely, the Sync-D metric evaluates the temporal consistency of dynamic lip movements, where lower values denote improved motion fidelity. 
Finally, the Expression-FID (E-FID) quantifies expression synchronization differences between generated content and ground truth videos, providing a quantitative assessment of expression accuracy.

\begin{figure}[!t]
    \centering
    \includegraphics[width=1.0\linewidth]{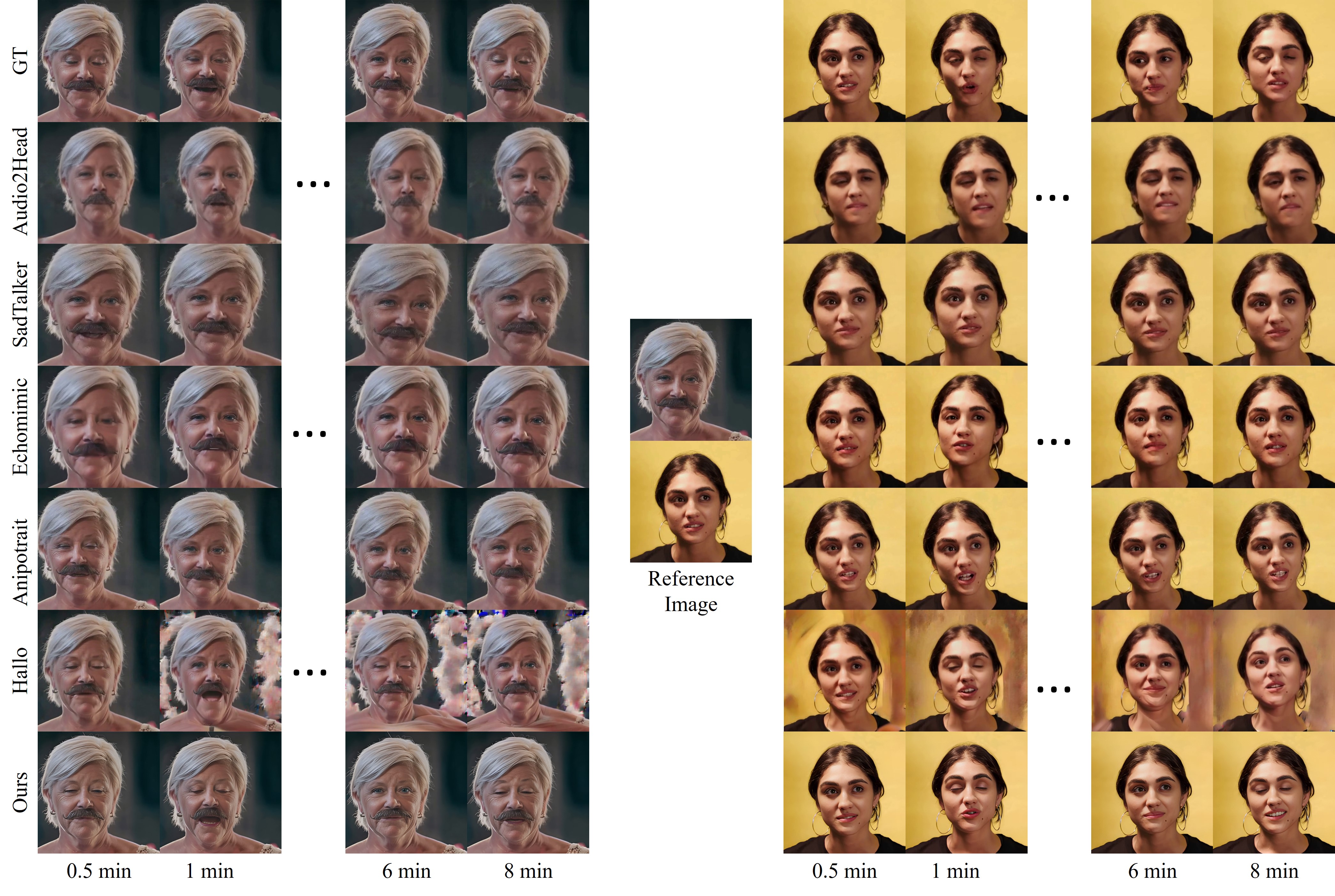}%comparison_celebv.png
    \vspace{-4mm}
    \caption{Qualitative comparison with exited approaches on CelebV data-set.}
    \label{fig:celebv} % 创建标签以便引用
\end{figure}

\begin{table}[!t]
    \centering
    \begin{tabular}{c|c|c|c|c|c}
        \toprule
        \textbf{Method} & \textbf{FID$\downarrow$} & \textbf{FVD$\downarrow$} & \textbf{Sync-C$\uparrow$} & \textbf{Sync-D$\downarrow$} & \textbf{E-FID$\downarrow$} \\ 
        \midrule
        Audio2Head  &57.879  &495.421  &\textbf{7.069}  &\textbf{7.916}  &60.538 \\
        SadTalker  &41.852  &588.173  &7.026  &7.931  &21.806  \\
        EchoMimic  &60.252  &805.067  &5.499  &9.482  &19.680  \\ 
        AniPortrait &49.626  &583.709  &3.810  &10.930  &22.220  \\ 
        Hallo  &82.715  &1088.158  &6.683  &8.420  &15.616  \\ 
        Ours &\textbf{37.944}  &\textbf{477.412}  &6.928  &8.307  &\textbf{14.682}  \\
        \midrule
        Real video & - & - &7.109 &7.938 & - \\ 
        \bottomrule
    \end{tabular}
    \vspace{-2mm}
    \caption{The quantitative comparisons with existed portrait image animation approaches on the CelebV data-set. }
    \vspace{-2mm}
\label{tab:celebv_results}
\end{table}

\begin{table}[!t]
    \centering
    \begin{tabular}{c|c|c|c|c|c}
        \toprule
        \textbf{Method} & \textbf{FID$\downarrow$} & \textbf{FVD$\downarrow$} & \textbf{Sync-C$\uparrow$} & \textbf{Sync-D$\downarrow$} & \textbf{E-FID$\downarrow$} \\ 
        \midrule
        Audio2Head  &50.449  &448.695  &6.269  &8.325  &38.981 \\
        SadTalker  &24.600  &380.866  &6.384  &8.169  &44.596  \\
        EchoMimic  &50.994  &854.826  &5.082  &9.675  &35.806  \\ 
        AniPortrait &24.301  &344.000  &3.975  &10.171  &41.307  \\ 
        Hallo  &28.186  &571.991  &6.610  &8.181  &36.793  \\ 
        Ours &\textbf{24.072}  &\textbf{360.192}  &\textbf{6.760}  &\textbf{8.156}  &\textbf{33.316}  \\
        \midrule
        Real video & - & - &7.088 &7.726 & - \\ 
        \bottomrule
    \end{tabular}
    \vspace{-2mm}
    \caption{The quantitative comparisons with existed approaches on the proposed ``Wild'' data-set.}
\label{tab:wild_results}
\end{table}

\begin{figure}[!t]
    \centering
    \includegraphics[width=1.0\linewidth]{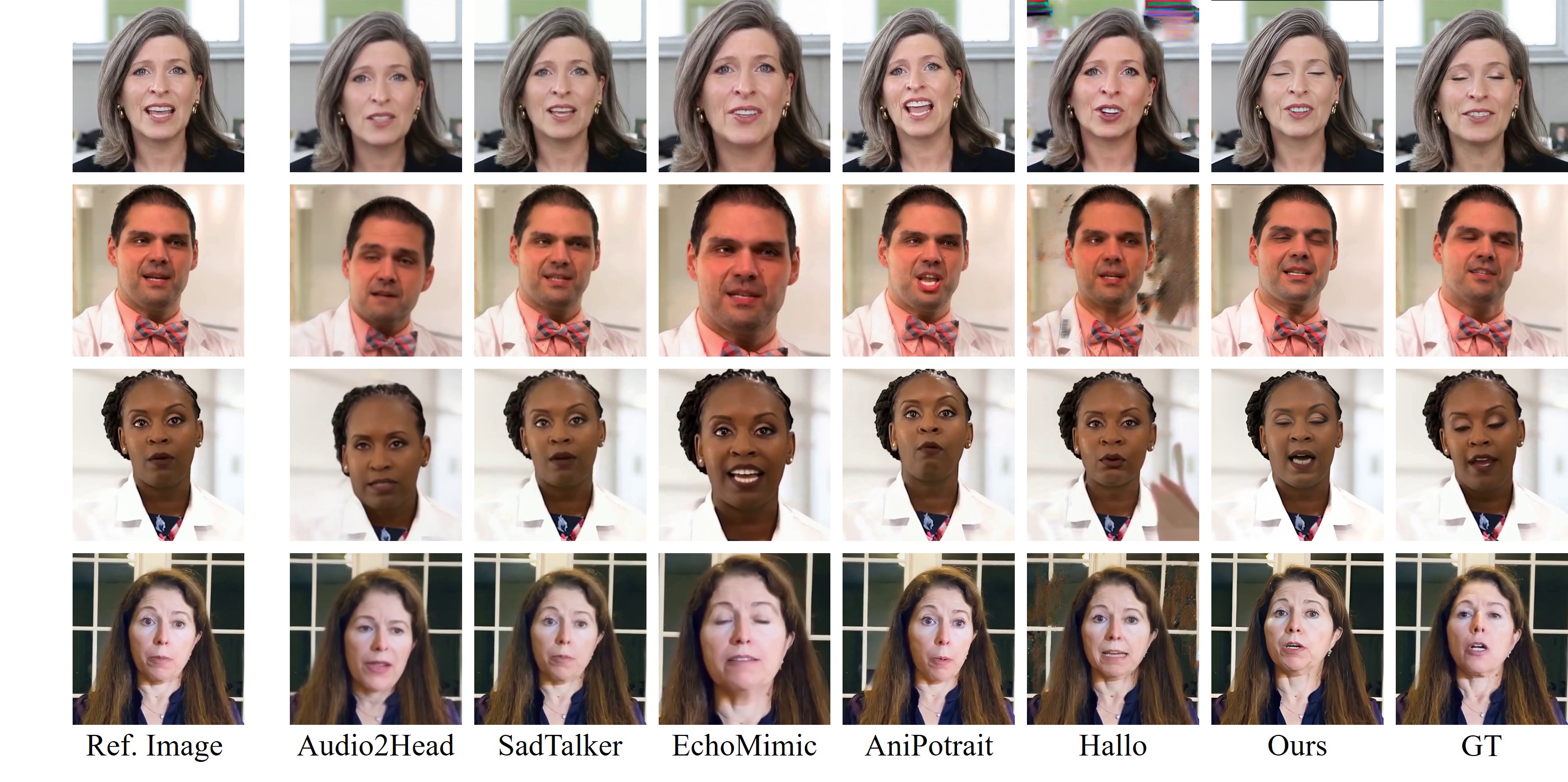}
    \vspace{-4mm}
    \caption{The qualitative comparison with existed approaches on the proposed ``Wild'' data-set.}
    \label{fig:wild} % 创建标签以便引用
    \vspace{-4mm}
\end{figure}

\begin{figure}[!t]
    \centering
    \includegraphics[width=1.0\linewidth]{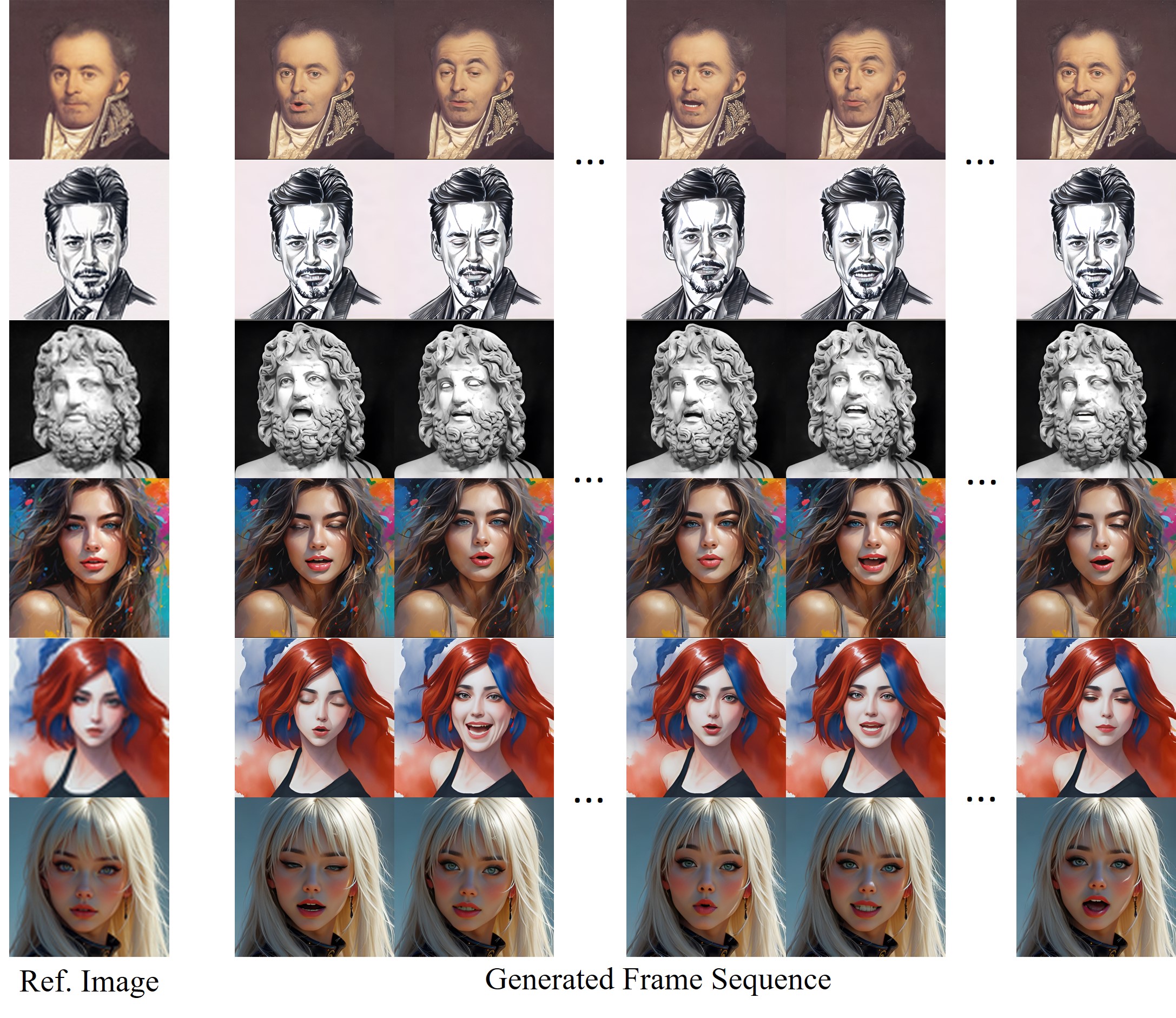} %styles.png
    \vspace{-6mm}
    \caption{Portrait image animation results given different portrait styles.}
    \vspace{-2mm}
    \label{fig:portrait_style} % 创建标签以便引用
\end{figure}

\textbf{Baseline Approaches.} 
We evaluate our framework against leading state-of-the-art techniques, including both non-diffusion and diffusion-based models. 
Non-diffusion models, such as Audio2Head and SadTalker, are compared with diffusion-based counterparts like EchoMimic, AniPortrait, and Hallo. 
Notably, EchoMimic and AniPortrait employ a parallel generation approach for long-duration outputs, while Hallo utilizes an incremental formulation. 
Unlike previous studies that focused on short-duration videos of only a few seconds, our evaluation is conducted on generated videos lasting 4 minutes, using looped audio from the benchmark dataset as the driving audio.  
To ensure a fair comparison, we have excluded the high-resolution enhancement module, maintaining the same output video resolution (512 × 512 pixels) as the existed approaches across all quantitative comparisons.

\subsection{Comparison with State-of-the-Art}
\textbf{Comparison on HDTF Dataset.} 
Table~\ref{tab:hdtf_results} and Figure~\ref{fig:hdtf} present quantitative and qualitative comparisons on the HDTF dataset. 
Our framework achieves the lowest FID of 16.616 and an E-FID of 6.702, demonstrating superior fidelity and perceptual quality. Additionally, our synchronization metrics, Sync-C (7.379) and Sync-D (7.697), further validate the effectiveness of our method. 
As illustrated in Figure~\ref{fig:fid_curve}, the extended inference duration significantly impacts FID metrics in existing diffusion-based approaches, leading to notable declines compared to their short-duration performance. 
In terms of lip and expression motion synchronization, parallel methods such as EchoMimic and AniPortrait exhibit marked deterioration. 
In contrast, our extended approach consistently demonstrates superior and stable performance across image and video quality, as well as motion synchronization, even as inference time increases.

\textbf{Comparison on CelebV Dataset.} 
Table~\ref{tab:celebv_results} and Figure~\ref{fig:celebv} present the quantitative and qualitative comparisons for the CelebV dataset. 
Our method achieves the lowest FID of 37.944 and an E-FID of 14.682, indicating superior animation quality. The FVD metric is reported at 477.412, suggesting a coherent video structure. 
Additionally, our Sync-C score of 6.928 demonstrates competitive performance relative to real video standards. 
Notably, the increased inference duration has resulted in a significant deterioration in both FID and FVD scores among existing methods, particularly with EchoMimic and Hallo, which exhibit marked degradation in FVD metrics. 
Additionally, Aniportrait demonstrates notable declines in lip synchronization and expression metrics.

\textbf{Comparison on the Proposed ``Wild'' Dataset.} 
Table~\ref{tab:wild_results} and Figure~\ref{fig:wild} offers additional quantitative and qualitative comparison results of the introduced ``Wild'' dataset.
Our method achieves an FID of 24.072 and an E-FID of 33.316, both indicative of high image quality. 
We also register a Sync-C score of 6.760 and a Sync-D of 8.156, alongside the highest FVD of 360.192, demonstrating superior coherent video structure. 

\textbf{Animation of Different Portrait Styles.}
Figure~\ref{fig:portrait_style}. This figure illustrates that our method is capable of processing a wide range of input types, including oil paintings, anime images, and portraits from generative models. 
These findings highlight the versatility and effectiveness of our approach in accommodating different artistic styles.

% Table~\ref{tab:wild_results} presents a quantitative comparison of various portrait image animation methods evaluated on the Wild dataset. Our proposed method exhibits superior performance relative to several existing techniques, achieving a Frechet Inception Distance (FID) score of 24.072 and an Enhanced Frechet Inception Distance (E-FID) score of 33.316, both of which indicate high image quality. Additionally, our approach attains a Synchronization Coefficient (Sync-C) score of 6.760, demonstrating effective synchronization of facial movements, while maintaining a competitive Synchronization Distance (Sync-D) score of 8.156. Notably, our method registers the highest Fréchet Video Distance (FVD) of 360.192, reflecting a coherent video structure in comparison to the other methods evaluated. Collectively, these results underscore the effectiveness of our method in delivering high-quality portrait image animations. Furthermore, Figure~\ref{fig:wild} offers additional qualitative comparison results to complement the quantitative findings.

% \begin{figure}[!t]
%     \centering
%     \includegraphics[width=1.0\linewidth]{fig/attn.png}
%     \caption{Visualization of the self-attention about reference image and the temporal attention about motion frames.}
%     \label{fig:attn} % 创建标签以便引用
% \end{figure}

\begin{table*}[t]
    \begin{minipage}{0.47\textwidth}
            \renewcommand{\arraystretch}{1.2}
    \makeatletter\def\@captype{table}\makeatother
            \resizebox{1\textwidth}{!}{
             \setlength{\tabcolsep}{0.5mm}{
     \begin{tabular}{c|c|c|c|c}
        \toprule
        \textbf{patch size} & \textbf{FID$\downarrow$} & \textbf{FVD$\downarrow$} & \textbf{Sync-C$\uparrow$} & \textbf{Sync-D$\downarrow$} \\ 
        \midrule
        0  &82.715 &1088.158 &6.683  &8.420  \\
        1  &\textbf{38.518}  &\textbf{491.338}  &\textbf{6.766}  &\textbf{8.387}  \\ 
        4  &39.615  &504.287  &6.712  &8.411   \\ 
        16 &44.172  &756.517  &6.431  &8.517    \\  
        \bottomrule
    \end{tabular}
            }}
    \caption{Quantitative comparison on the CelebV dataset given different patch sizes of patch drop augmentation. A patch size of 0 indicates no patch drop.} 
    \label{tab:ablation_patch_rate}
    
  \end{minipage}\quad
  \begin{minipage}{0.47\textwidth}
            \renewcommand{\arraystretch}{1.2}

    \makeatletter\def\@captype{table}\makeatother
            \resizebox{1\textwidth}{!}{
            \setlength{\tabcolsep}{0.5mm}{
    \begin{tabular}{c|c|c|c|c}
        \toprule
        \textbf{Drop rate} & \textbf{FID$\downarrow$} & \textbf{FVD$\downarrow$} & \textbf{Sync-C$\uparrow$} & \textbf{Sync-D$\downarrow$} \\ 
        \midrule
         0   &82.715 &1088.158 &6.683 &8.420\\
        0.1  &41.687  &535.212  &6.692  &8.395  \\ 
        0.25  &\textbf{38.518}  &\textbf{491.338}  &\textbf{6.766}  &\textbf{8.387}   \\ 
        0.5  &39.642  &513.314  &6.687  &8.515    \\  
        \bottomrule
    \end{tabular}
            }}
            \caption{Quantitative comparison on the CelebV dataset given different drop rate of patch drop augmentation. A drop rate of 0 indicates no patch drop.} 
            \label{tab:ablation_mask_rate}
  \end{minipage}
  \vspace{-2mm}
\end{table*}

\begin{table}[!t]
    \centering
    \begin{tabular}{c c|c|c|c|c}
        \toprule
        \textbf{Gaussian noise}& \textbf{Patch drop} & \textbf{FID$\downarrow$} & \textbf{FVD$\downarrow$} & \textbf{Sync-C$\uparrow$} & \textbf{Sync-D$\downarrow$} \\ 
        \midrule
        & &82.715  &1088.158  &6.683 &8.420 \\ 
        \checkmark & & 78.283 & 984.876 & 6.701 & 8.415 \\ 
        &\checkmark  & 38.518 & 491.338 & 6.766 &8.387   \\ 
        \checkmark &\checkmark  &\textbf{37.944} &\textbf{477.412} & \textbf{6.928} &  \textbf{8.307}  \\  
        \bottomrule
    \end{tabular}
    \vspace{-2mm}
    \caption{Ablation study of the patch drop and Gaussian noise augmentation on the CelebV data-set.}
\label{tab:ablation_combination}
    \vspace{-2mm}
\end{table}

\subsection{Ablation Studies}
\textbf{Different Patch Drop Size.}
Table~\ref{tab:ablation_patch_rate} illustrates the effects of varying patch drop sizes on performance metrics. 
A patch size of 0 signifies no patch drop, while our implementation employs a patch size of 1. 
The results indicate that patch drops enhance visual outcomes, as evidenced by improvements in FID and FVD, and contribute to a degree of enhancement in motion synchronization capabilities.

% As shown in Table~\ref{tab:ablation_patch_rate}, the impact of varying patch drop sizes on performance metrics is summarized in Table \ref{tab:ablation_patch_rate}. Notably, a patch size of \( s=1 \) achieves the lowest Fréchet Inception Distance (FID) of \( 36.053 \) and the lowest Fréchet Video Distance (FVD) of \( 478.406 \), indicating superior quality. In contrast, an \( s=0 \) configuration results in the highest FID and FVD values. 
% While the Synchronization Consistency (Sync-C) peaks at \( 6.081 \) with a patch size of \( s=1 \), the lowest Sync-D (synchronization discrepancy) is recorded at \( 8.663 \) for the same configuration. 
% These findings suggest that optimizing patch size can significantly enhance the quality and consistency of outputs.

\begin{figure}[!t]
 	\begin{minipage}{0.45\textwidth}
        \centering
    \includegraphics[width=1.0\textwidth]{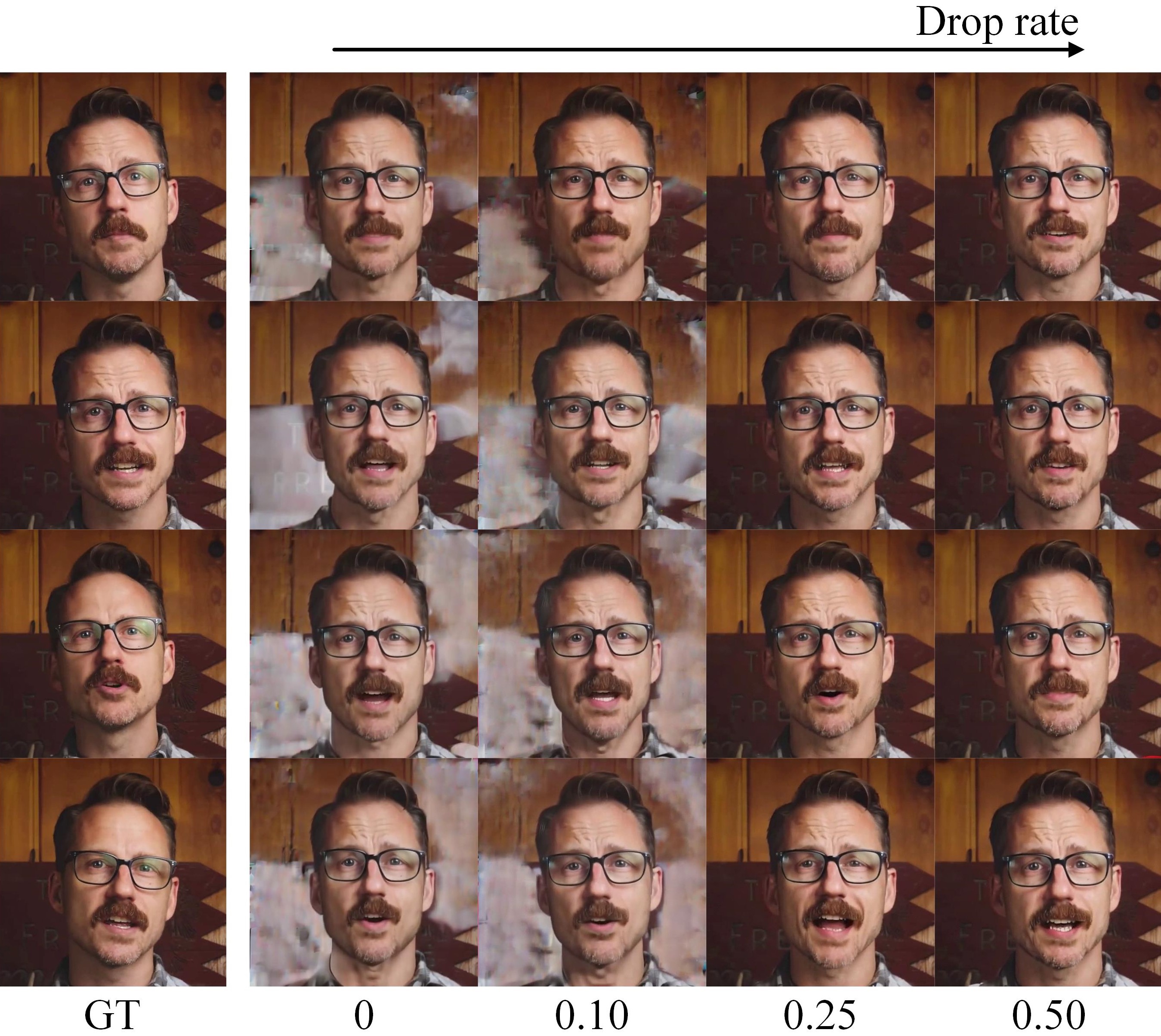} 
    \vspace{-2mm}
    \caption{Qualitative comparison of different patch drop rate applied to motion frames on the CelebV data-set.}
    \vspace{-2mm}
    \label{fig:ablation_mask_rate}
	\end{minipage}
 \hfill
 \begin{minipage}{0.49\textwidth}
    \centering
      \includegraphics[width=1.0\textwidth]{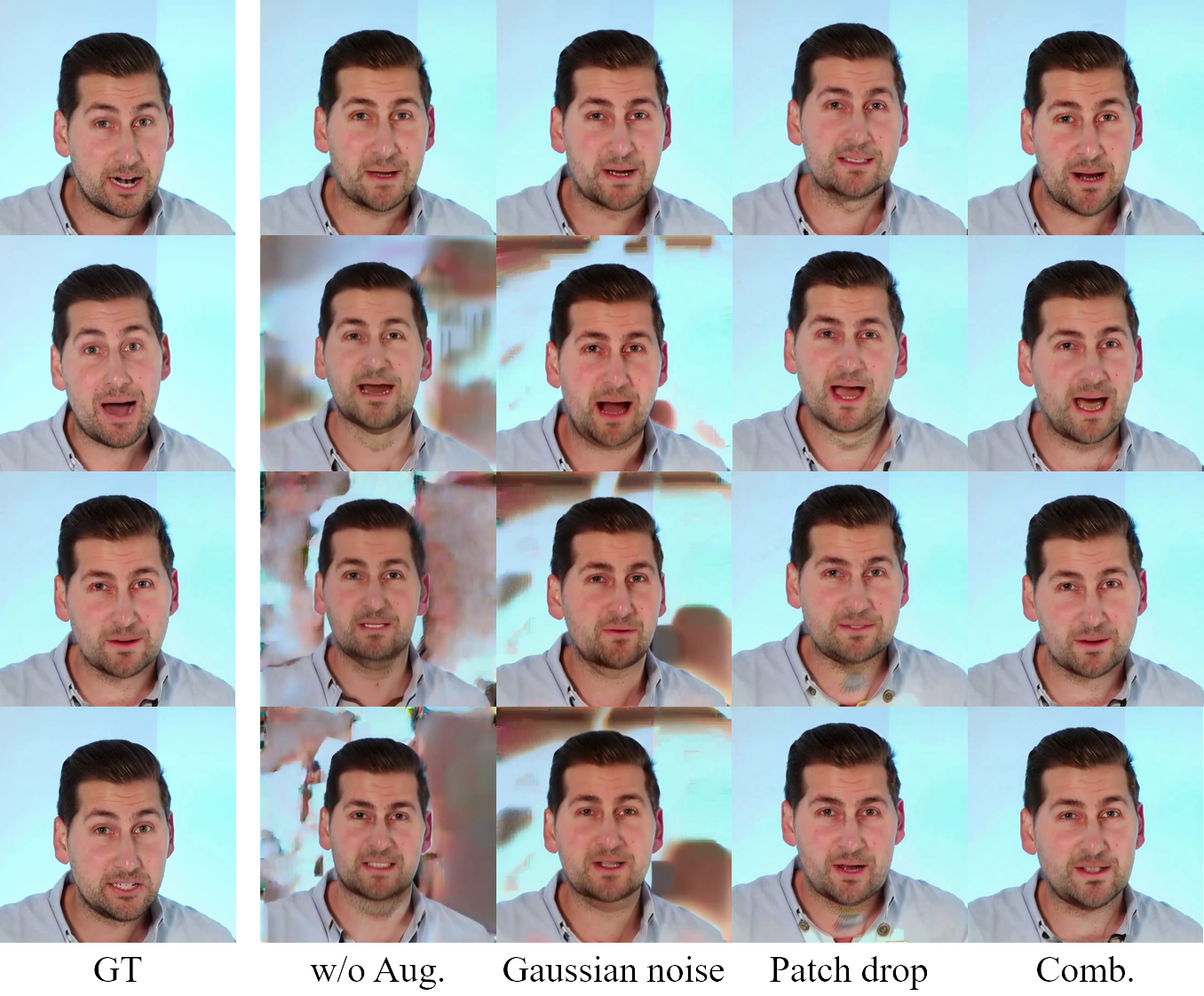}
      \vspace{-2mm}
      \caption{Qualitative ablation study of the patch drop, Gaussian noise augmentation and combination of both approaches.}\label{fig:ablation_combination}
      \vspace{-2mm}
\end{minipage}
\end{figure}

\textbf{Different Patch Drop Rate.} Table~\ref{tab:ablation_mask_rate} and Figure~\ref{fig:ablation_mask_rate} present a comparative analysis of varying drop rates applied to motion frames. A drop rate of 0.25 achieves the lowest FID score of 38.518 and FVD of 491.338, indicating improved image quality and coherence. 

% Table~\ref{tab:ablation_mask_rate} presents a comparative analysis of the impact of varying drop rates applied to motion frames on key performance metrics. 
% As observed, a drop rate of 0.25 yields the lowest Frechet Inception Distance (FID) score of 36.746 and a corresponding Fréchet Video Distance (FVD) of 486.515, indicating improved image quality and coherence in animation. 
% Furthermore, this drop rate achieves a Synchronization Coefficient (Sync-C) of 6.313, demonstrating effective synchronization of facial movements. 
% In contrast, a higher drop rate of 0.5 results in a slight increase in FID (39.670) and FVD (516.423), alongside a decrease in Sync-C (5.829), suggesting that excessive frame dropping may negatively impact synchronization. 
% The results collectively highlight the significance of optimizing the drop rate to enhance the overall performance of portrait image animations.

\textbf{Effectiveness of Augmentation Strategies.} 
Table~\ref{tab:ablation_combination} and Figure~\ref{fig:ablation_combination} evaluate different augmentation strategies. Gaussian noise alone results in a high FID of 82.715 and FVD of 1088.158, indicating suboptimal quality. 
The patch drop strategy significantly improves these metrics, reducing FID to 38.518 and FVD to 491.338. Notably, the combined strategy further enhances performance, achieving the lowest FID of 37.944 and FVD of 477.412, alongside the highest Sync-C score of 6.928. 
Thus, the combined augmentation method proves to be the most effective in generating high-quality motion frames.

\begin{figure}[!t]
    \centering
    \includegraphics[width=1.0\linewidth]{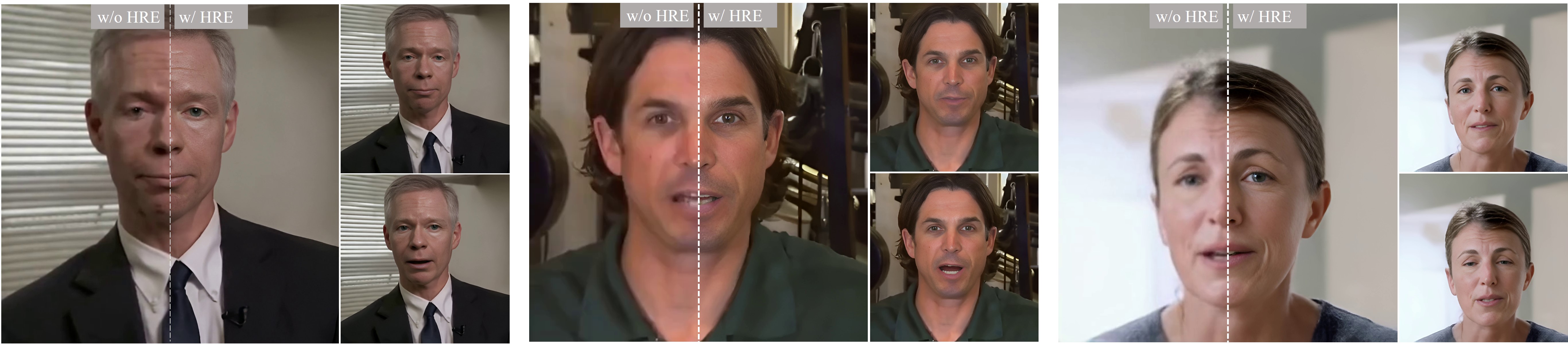}
    \vspace{-2mm}
    \caption{Qualitative comparison of the portrait image animation results with and without high-resolution enhancement.}
    \vspace{-2mm}
    \label{fig:ablation_super_resolution} % 创建标签以便引用
\end{figure}

\textbf{Effectiveness of High-Resolution Enhancement.} 
The effectiveness of high-resolution enhancement techniques is illustrated in Figure~\ref{fig:ablation_super_resolution}, which demonstrates improved animation quality via video super-resolution. 

\begin{figure}[!t]
    \centering
    \includegraphics[width=0.8\linewidth]{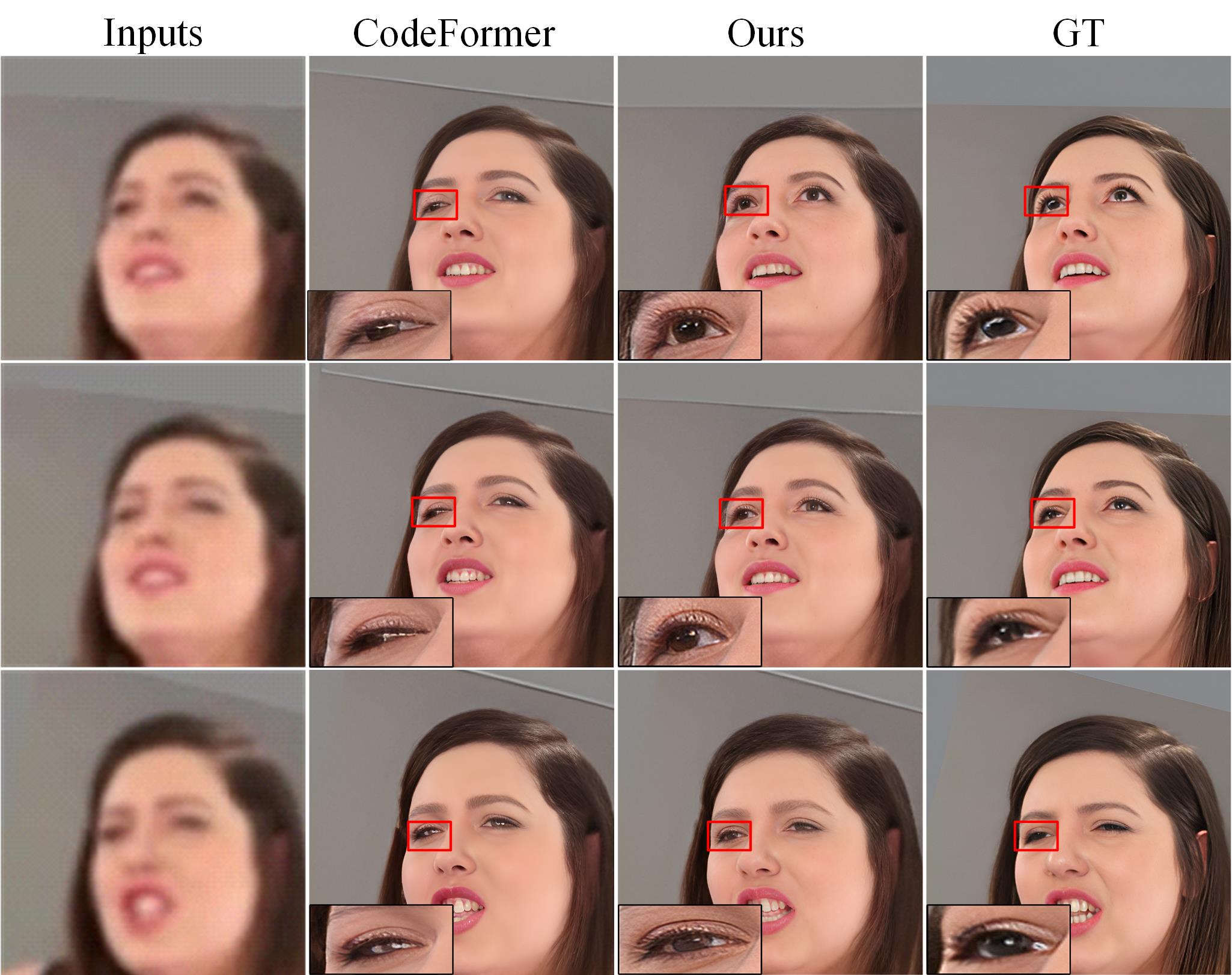}\vspace{-2mm}
    \caption{Qualitative comparison between different high-resolution enhancement methods.}
    \vspace{-2mm}
    \label{fig:ablation_temp_align} % 创建标签以便引用
\end{figure}

\begin{figure}[!t]
    \centering
    \includegraphics[width=1.0\linewidth]{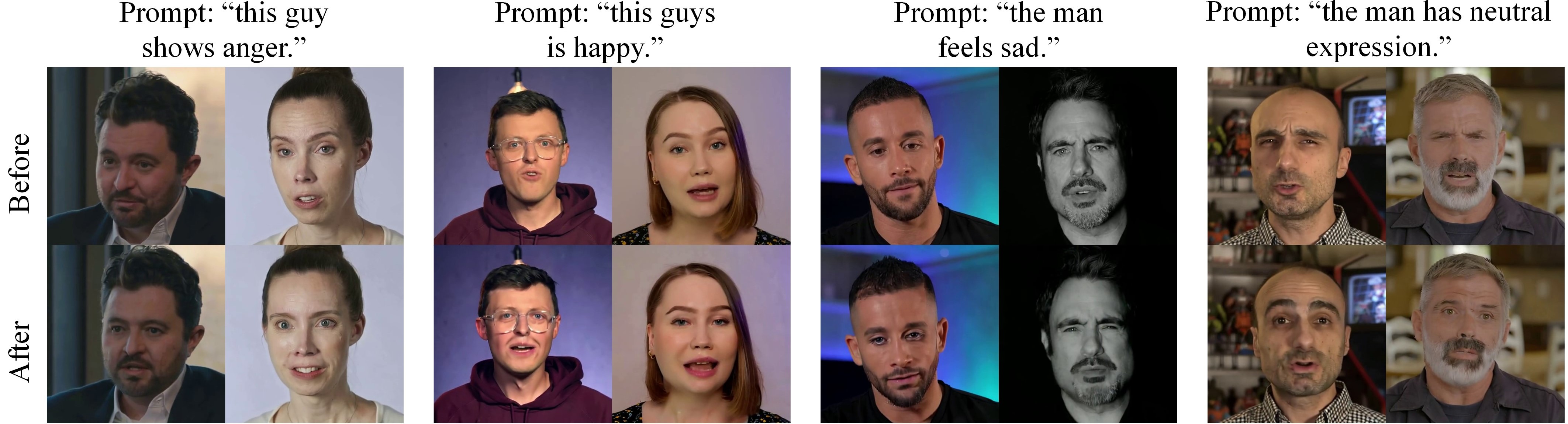}
    \vspace{-6mm}
    \caption{Qualitative comparison of portrait animation before and after applying the textual prompts.}
    \label{fig:ablation_emotion_control} % 创建标签以便引用
    \vspace{-2mm}
\end{figure}

\begin{figure*}[!t]
\centering
\subfigure[Reference image]{
\begin{minipage}[b]{0.47\textwidth}\vspace{-2mm}
\includegraphics[width=1.0\textwidth]{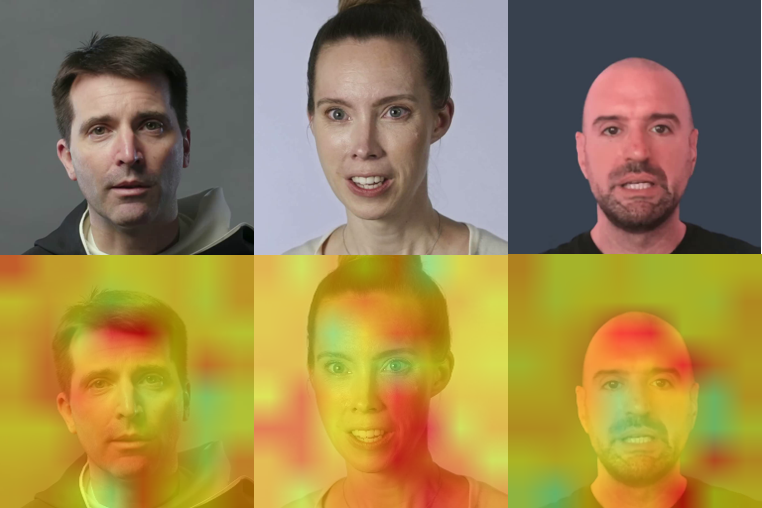}
\label{fig:att_ref}
\end{minipage}
}
\hspace{1mm}
\subfigure[Motion frames]{
\begin{minipage}[b]{0.47\textwidth}\vspace{-2mm}
\includegraphics[width=1.0\textwidth]{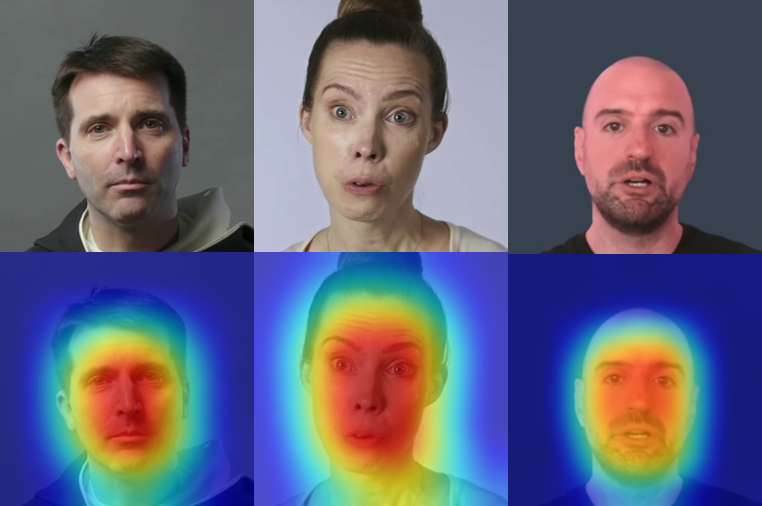}
\label{fig:att_mot}
\end{minipage}
}
\vspace{-2mm}
\caption{Attention map visualization of the reference image and motion frames.}
\vspace{-2mm}
\label{fig:att_map}
\end{figure*}

\textbf{Comparison between Different High-Resolution Enhancement Methods.}
Figure~\ref{fig:ablation_temp_align} provides a qualitative comparison of other image-based enhancement methods. 
The analysis reveals that integrating super-resolution with temporal alignment significantly enhances visual fidelity, reduces artifacts, and increases image sharpness, resulting in a more coherent and realistic representation of facial features and expressions.

\textbf{Effectiveness of Textual Prompt.} 
The integration of textual prompts into our portrait image animation framework significantly enhances the control over generated animations, as illustrated in Figure~\ref{fig:ablation_emotion_control}. 
The comparative analysis demonstrates that textual prompts facilitate precise manipulation of facial expressions and emotional nuances, allowing for a more tailored animation output. 
By providing explicit instructions regarding desired emotional states, the model exhibits improved responsiveness in generating animations that align closely with the specified prompts. 

\textbf{Attention Map Visualization.}
Figure~\ref{fig:att_map} presents the attention map visualization, which highlights both the reference image and the temporal attention associated with the motion frames. 
The results indicate that the reference image indeed influences the overall appearance of the portrait and background due to the implementation of patch drop augmentation. 
In contrast, the motion frames predominantly focus on regions related to facial motion, underscoring their role in capturing dynamic attributes in the generated animation.

\subsection{Limitations and Future Work}
Our method for long-duration, high-resolution portrait image animation has several limitations. 
(1)~Reliance on a single reference image constrains the diversity of generated expressions and poses, indicating a need for multiple references or advanced models capable of synthesizing varied facial features. 
(2)~While the patch-drop data augmentation technique effectively preserves motion dynamics, it may introduce artifacts; thus, future research should investigate alternative strategies or adaptive mechanisms for content-specific corruption. 
(3)~The substantial computational demands of generating 4K resolution videos necessitate optimization and hardware acceleration to enable real-time applications.
\section{Conclusion}
This paper presents advancements in portrait image animation through the enhanced capabilities of the Hallo framework. 
By extending animation durations to tens of minutes while maintaining high-resolution 4K output, our approach addresses significant limitations of existing methods. 
Specifically, innovative data augmentation techniques, including patch-drop and Gaussian noise, ensure robust identity consistency and reduce appearance contamination.
Furthermore, we implement vector quantization of latent codes and employ temporal alignment techniques to achieve temporally consistent 4K videos. 
Additionally, the integration of audio-driven signals with adjustable semantic textual prompts enables precise control over facial expressions and motion dynamics, resulting in lifelike and expressive animations. 
Comprehensive experiments conducted on publicly available datasets validate the effectiveness of our method, representing a significant contribution to the field of long-duration, high-resolution portrait image animation.

\bibliography{iclr2025_conference}
\bibliographystyle{iclr2025_conference}

\end{document}